%% file: MS.tex
\documentclass[journal,twoside,web]{ieeecolor}

\usepackage{xcolor}

\newcounter{sideremark}

\usepackage{generic}

\usepackage{soul}
\usepackage{graphicx,color}
\usepackage{algorithm}
\usepackage{algpseudocode}
\usepackage{setspace}
\usepackage{mwe}
\usepackage{multicol}
\usepackage[utf8]{inputenc}
\usepackage[english]{babel}

\usepackage{amssymb,amsmath}

\usepackage{amsfonts}
\usepackage{amssymb}
\usepackage{scalefnt}
\usepackage{xr}
\usepackage{flushend}
\usepackage{enumerate}
\usepackage{bbm}
\usepackage{dsfont}
\usepackage{comment}
\usepackage{cite}
\usepackage{amsthm}

\newtheorem{theorem}{Theorem}
\newtheorem{proposition}{Proposition}
\newtheorem{corollary}{Corollary}%[theorem]
\newtheorem{lemma}{Lemma}

\newtheorem{assumption}{Assumption}
\theoremstyle{definition}
\newtheorem{definition}{Definition}[section]

\usepackage{dsfont}
\allowdisplaybreaks
\usepackage[subtle,tracking=normal]{savetrees}

\usepackage{hyperref}
\hypersetup{hidelinks=true}
\usepackage{textcomp}
\def\BibTeX{{\rm B\kern-.05em{\sc i\kern-.025em b}\kern-.08em
    T\kern-.1667em\lower.7ex\hbox{E}\kern-.125emX}}
\begin{document}
\title{Resilient Distributed Optimization for \\
Multi-Agent Cyberphysical Systems}
\author{Michal Yemini, Angelia Nedi\'c, Andrea J. Goldsmith, Stephanie Gil%
 \thanks{
  M.~Yemini is with the Faculty of Engineering, Bar-Ilan University, Israel ({\tt\small michal.yemini@biu.ac.il}).
 A.~Nedi\'{c} is with the School of Electrical, Computer and Energy Engineering, Arizona State University, Tempe, AZ 85281 USA ({\tt\small Angelia.Nedich@asu.edu}).
 A.~J.~Goldsmith is with the Department of Electrical and Computer Engineering, Princeton University, Princeton, NJ 08544 USA ({\tt\small goldsmith@princeton.edu}).
 S.~Gil is with the Computer Science Department, School of Engineering and Applied Sciences, Harvard University, Cambridge, MA 02139 USA ({\tt\small sgil@seas.harvard.edu}).
}
\thanks{
This work was partially supported by NSF awards \#CNS-2147631,  \#CNS-2147641 and \#CNS-2147694.  M.~Yemini and A.~J.~Goldsmith were partially supported by  AFOSR award \#002484665.}
\thanks{This paper was presented in part at the IEEE Conf.~Decis.~Control, \cite{accepted_cdc_us}.}
\thanks{
\vspace{0.1cm}
\hspace{-0.45cm}
\noindent\fbox{%
    \parbox{0.48\textwidth}{%
© 2025 IEEE. Personal use of this material is permitted. Permission from IEEE must be
obtained for all other uses, in any current or future media, including
reprinting/republishing this material for advertising or promotional purposes, creating new
collective works, for resale or redistribution to servers or lists, or reuse of any copyrighted
component of this work in other works.}
}}
}
\input{commands}

\maketitle

%%%%%%%%%%%%%%%%%%%%%%%%%%%%%%%%%%%%%%%%%%%%%%%%%%%%%%%%%%%%%%%%%%%%%%%%%%%%%%%%
\begin{abstract}

This work focuses on the problem of distributed optimization in multi-agent cyberphysical systems, where a legitimate agent's iterates are influenced both by the values it receives from potentially malicious neighboring agents, and by its own self-serving target function. We develop a new algorithmic and analytical framework to achieve resilience for the class of problems where stochastic values of trust between agents exist and can be exploited. In this case, we show that convergence to the true global optimal point can be recovered, both in mean and almost surely, even in the presence of malicious agents. Furthermore, we provide expected convergence rate guarantees in the form of upper bounds on the expected squared distance to the optimal value. Finally, numerical results are presented that validate our analytical convergence guarantees even when the malicious agents compose the majority of agents in the network and where existing methods fail to converge to the optimal nominal points.
\end{abstract}

%%%%%%%%%%%%%%%%%%%%%%%%%%%%%%%%%%%%%%%%%%%%%%%%%%%%%%%%%%%%%%%%%%%%%%%%%%%%%%%%
\vspace{-0.3cm}
\section{Introduction}

Distributed optimization is at the core of various multi-agent tasks including distributed control and estimation, multi-robot tasks such as mapping, and many learning tasks such as Federated Learning~\cite{macDistOpt,martinezDistributedText,fedLearning}. Owing to a long history and much attention in the research community, the theory for distributed optimization has matured, leading to several important results that provide rigorous performance guarantees in the form of convergence and convergence rate for different function types, underlying graph topologies, and noise~\cite{distr-subgrad2009,boydDistOpt,dist_opt_self_directed_time,Lan-asynch,Lan-sampling}. However, in the presence of malicious activity, many of these known results are no longer applicable, requiring a new theoretical characterization of performance for the adversarial case.

The growing prevalence of multi-agent cyberphysical systems, and their reliance on distributed optimization methods for correct functioning in the real world, underscores the criticality of understanding the vulnerabilities of these systems. In particular, malicious agents can greatly interfere with the result of a distributed optimization scheme, driving the convergence to a non-optimal solution or preventing convergence altogether. They can accomplish this by either not sharing key information or by manipulating key information such as the shared gradients, which are critical for the correct functioning of the distributed optimization scheme~\cite{Ravi_distributed_opt_icassp,Sundaram_distributed_opt_malicious_2019,stateEstMalAgents}. Note that while well-established stochastic optimization methods characterize the effect of noise in distributed multi-agent  systems~\cite{Tsitsiklis_1986,Angelia_Stochastic_gradient_Push_2016}, malicious agents have the ability to inject intentionally biased or manipulated information which can lead to a greater potential damage for these systems. As a result, recent works have increasingly turned their attention to the investigation of robust and resilient versions of distributed optimization methods in the face of malicious intent and/or severe (potentially biased) noise~\cite{Ravi_distributed_opt_icassp,Sundaram_distributed_opt_malicious_2019,stateEstMalAgents,soniaAdv,saulnier2017resilient}. These approaches can be coarsely divided into two categories: those that use the transmitted data between nodes to infer the presence of anomalies (for example see~\cite{bulloUnreliable,Sundaram_distributed_opt_malicious_2019}), and those that exploit additional side information from the network or the physicality of the underlying cyberphysical system to provide additional channels of resilience~\cite{CPS_Security_sastryBook,sastryCPS_survivability,SecureArray}.

We are interested in investigating the class of problems where the physicality of the system plays an important role in achieving new possibilities of resilience for these systems.  Indeed, the physicality of cyberphysical systems has been shown to provide many new channels of verification and establishing \emph{inter-agent trust} through watermarking~\cite{Sinopoli_Watermark}, wireless signal characteristics~\cite{SecureArray,AURO}, side information~\cite{shannonSideInfo}, and camera or LiDAR data cross-validation~\cite{henrik2014trust}. By exploiting these physicality-based measurements,  agents can extract additional information about the trustworthiness of their neighbors.

We capitalize on this observation, which motivates us to focus on a class of problems where the existence of additional information in the system can be exploited to arrive at much stronger performance results in the adversarial case -- what we refer to as \emph{resilience}.  We abstract this information as a value $\alpha_{ij}$ that indicates the likelihood with which an agent $i$ can trust data received from another agent $j$. We show that under mild assumptions, when this information is available, several powerful results for distributed optimization can be recovered such as: 1) \emph{\textbf{convergence to the true optimal point}} in the case of minimizing the sum of strongly convex functions, and 2) \emph{\textbf{characterization of convergence rate}} that depends on the network topology, the amount of trust observations acquired, and the number of legitimate and malicious agents in the system.

\vspace{-0.25cm}
\subsection{Related work}
In the absence of malicious agents, the legitimate agents can construct iterates converging to an optimal point $x_{\mathcal{L}}^{\star}$ by using either their gradients, or sub-gradients when their objective functions are not differentiable. Each agent $i$ updates its data value by considering the data values of its neighbors, and its self-serving gradient direction of its objective function $f_i$ or the directions obtained from its neighbors. Convergence to an optimal point $x_{\mathcal{L}}^{\star}$ can be achieved for constrained multi-agent problems
in~\cite{distr-subgrad2009,nedicConstrainedConsensus, ram_distributed_stochastic_subgradient_2010, Duchi2012,dist_opt_self_convex_undirected1,dist_opt_self_convex_undirected2,dist_opt_self_directed_time,dist_opt_self_strongly_undirected,Angelia_Stochastic_gradient_Push_2016} and with limited gradient information~\cite{Lina-limitedcommgradmethods,Limited_Rajarshi_2022}. 
Additionally, a zero-order method has been proposed in~\cite{Lina-zero-order}. 
    Some works, such as~\cite{distr-subgrad2009}, assume that the weight matrices, which dictate how agents incorporate the data they receive from their neighbors, are doubly-stochastic. However,
    works such as~\cite{dist_opt_self_convex_undirected1} overcome this assumption by performing additional weighted averaging steps.  
    Finally, it has been established that the convergence rate of distributed gradient algorithms with diminishing step size is at best $O(\frac{1}{T})$ where $T$ is the algorithm running time, see for example \cite{dist_opt_self_strongly_undirected}. 

 To harm the system, a malicious agent can send falsified data to their legitimate neighbors. If the legitimate agents are unaware that this data comes from malicious neighbors, 
then the malicious agents will succeed in controlling the system~\cite{Abbas_dist_opt_trusted_nodes,Baras_2019_trusted,Zhao_dist_opt_trusted_nodes,Ravi_distributed_opt_icassp,Sundaram_distributed_opt_malicious_2019,Turan_byzantine_central_controller,resilient_distributed_subgradient_2021,secure_clipping_common_minimizer_2023}.  
To combat the harmful effect of an attack, the approach taken in~\cite{Abbas_dist_opt_trusted_nodes,Baras_2019_trusted,Zhao_dist_opt_trusted_nodes} requires the pre-existence of a set of trusted agents such that all other agents (legitimate or malicious) are connected to at least one trusted agent. Nonetheless, this approach 
is unrealistic when communication is sporadic such as in robotic and ad-hoc networks. 
The approaches in~\cite{Ravi_distributed_opt_icassp,Sundaram_distributed_opt_malicious_2019,Ding_perturbation_state_gradient,Turan_byzantine_central_controller,resilient_distributed_subgradient_2021,secure_clipping_common_minimizer_2023}
rely on the agent data values to detect and discard malicious inputs.  
In general, data-based approaches
have an upper bound on the number of tolerable malicious agents which cannot exceed half of the network connectivity; in some cases, this condition can be relaxed to half of the number of agents in the system - see, for example, \cite{secure_clipping_common_minimizer_2023}. Thus, they are not robust when malicious agents form a majority~\cite{Turan_byzantine_central_controller}. 
When the number of malicious agents exceeds the tolerable number, the attack succeeds and malicious agents evade detection. In contrast with the existing works, our proposed method provides a significantly stronger resilience to malicious activity by
exploiting the physical aspect of the problem, i.e., the wireless medium. 
Thus, each legitimate agent can learn trustworthy neighbors while optimizing the system objective. Our prior work \cite{ourTRO} studies the implications of the agents' learning ability, with regards to the trustworthiness of their neighbors, on distributed \textit{consensus} systems. This work considers the more general case of distributed \textit{optimization} systems where the agent's goal is to minimize the sum of their local objective function under limited information exchange. 

Finally, this work also relates to stochastic optimization, see for example \cite{DuchiHaSi11-adaptive-sgd,Lan1,Lan2,var-red-johnson,afrooz-variablebatch,Wilson-seq-stoch,MDwithaveraging,ShiPu2021-tracking}. However, unlike the typical assumption that the stochastic gradients are unbiased and statistically independent of the weights, in this work the stochastic gradients are biased, where the bias occurs due to the adversarial inputs of the malicious agents.  Furthermore, learning  the trustworthiness of neighboring agents and adjusting agents' weight accordingly leads to a correlation between the agents' values and the weights that are assigned to them. To this end, our analysis could not rely on previous results when analyzing the rate of convergence of the agents' dynamic. 

\vspace{-0.25cm}
\subsection{Paper contribution}
This work extends our earlier paper \cite{accepted_cdc_us}, which formalized the problem of resilient distributed optimization with trust observations and presented an upper bound on the convergence rate in the mean squared sense. 
This work excluded the proofs for these theoretical results due to space limitations. The key contributions of this work beyond those of [1] are as follows: 
1)~We present the analysis of all the proofs that were omitted from \cite{accepted_cdc_us}.    
2)~We provide additional insights and discussions for the theoretical results that are included in \cite{accepted_cdc_us}. Specifically, this work includes extensive discussions regarding the assumptions that are used in this paper regarding the statistics of the trust observations and their relaxations.
3)~We extend the convergence in the second moment that follows from~\cite{accepted_cdc_us} and establish the almost sure convergence to the true optimal point. Additionally, we derive convergence in the $r$th mean for any $r>0$, both as a result of the almost sure convergence and by definition. These convergence results lead to an additional method to derive upper bounds on the convergence rate of the second moment which we present in this work. 
4)~We present additional numerical simulations that significantly extend the one-dimensional consensus with constraint setup to higher dimensional constrained optimization setups. Furthermore, we present in the supplementary material a comparison between the numerical simulations and the upper bounds we establish in this work.

\vspace{-0.25cm}
\subsection{Paper organization}
The rest of this paper is organized as follows: Section \ref{sec:problem_formulation} presents the system model and problem formulation. Section \ref{sec:algorithm} proposes our algorithm for resilient distributed optimization. Section \ref{sec:learning_trusted_neighbors} presents our learning mechanism for detecting malicious agents and 
provides upper bounds on the probability of misclassifying malicious and legitimate agents.
Sections \ref{sec:analytical_results_correct_class_time} 
and \ref{sec:finite_time_analysis},
respectively, present asymptotic and finite time regime convergence results. Finally, Section \ref{sec:numerical_results} presents numerical results that validate our analytical results, and Section \ref{sec:conclusions} concludes the paper.

\vspace{-0.2cm}

\section{Problem Formulation}\label{sec:problem_formulation}
We consider a multi-agent system of $n$ agents communicating over a network,
which is represented by an undirected graph, $\mathbb{G}=(\mathbb{V},\mathbb{E})$.
The node set $\mathbb{V}=\{1,\hdots,n\}$ 
represents the agents and the edge set $\mathbb{E}\subset\mathbb{V}\times\mathbb{V}$ 
represents the set of communication links, with $\{i,j\}\in\mathbb{E}$ indicating that agents $i$ and $j$ are connected.
We denote by $\NN_i$ the set of neighbors of agent $i$, that is 
$\NN_i\triangleq \{j:\{i,j\}\in\mathbb{E}\}$.

We study the case where an unknown subset of the agents is malicious and the trustworthy agents are learning which neighbors they can trust. Thus,  $\mathbb{V}=\LL\cup\MM$ where $\mathcal{L}$ is the nonempty and finite set of legitimate agents that  execute computational tasks and share their data truthfully,
while $\mathcal{M}$ denotes a finite set of agents that are not truthful.
{\it The sets $\mathcal{L}$ and $\mathcal{M}$ are defined solely for the sake of analysis, and none of the legitimate agents knows if it has malicious neighbors, or not, at any time.}
Throughout the paper, we will use
the subscripts $\LL$ and $\MM$ to denote the various quantities related to
legitimate and malicious agents, respectively.

 We are interested in a general distributed optimization problem, where the legitimate agents aim at optimizing a common objective whereas the malicious agents try to impair the legitimate agents by malicious injections of harmful data. 
The aim of the legitimate agents is to minimize distributively the sum of their objective functions over a constraint set $\XX\subset\mathbb{R}^d$, i.e.,
    \begin{flalign}\label{eq:dist_opt_obj}
   x_{\LL}^{\star}=\arg\min_{x\in\XX} f_{\mathcal{L}}(x),\hbox{ with }
   f_{\mathcal{L}}(x)=\frac{1}{|\LL|}\sum_{i\in\mathcal{L}}f_{i}(x),
     \end{flalign}
     with a unique minimizer.
     By choosing a local update rule and exchanging some information with their neighbors, the legitimate agents want to determine the optimal solution 
     $x_{\LL}^{\star}$ in~\eqref{eq:dist_opt_obj}. 
     In contrast, the malicious agents aim to
     either lead the legitimate agents to a common non-optimal value  $x\in\XX$ such that $f_{\mathcal{L}}(x)>f_{\mathcal{L}}(x^{\star})$,
     or prevent the convergence of an optimization method employed by the legitimate agents.
    \paragraph*{Notation}
        We let  $x^T$ denote the transpose of $x$, where $x\in\mathbb{R}^d$.
We denote 
by $\|x\|\triangleq \sqrt{x^Tx}$ the $\ell_2$ vector norm.
We let $\Pi_{\XX}(x)$ be the projection of  $x$ onto the set $\XX$, i.e.,
\[\Pi_{\XX}(x) = \argmin_{y\in\XX} \|y-x\|.\]
We use $\mathds{1}_{\{A\}}$ to denote the indicator function of the event $A$, which is equal to one if the event $A$ is true and zero otherwise.
Finally, we denote by $\EEop[\cdot]$ the expectation operator, and by $\EEop[\cdot|\cdot]$ the conditional expectation, where we arbitrarily define  $\EEop[\cdot|A]=0$ whenever the probability of the conditioned event $A$ is $0$.

%--------------------------
\vspace{-0.2cm}
\subsection{Trust values} \label{sec:trustVals}
We employ a probabilistic framework of trustworthiness where we assume the availability of stochastic \emph{observations of trust} between communicating agents.
This information is abstracted in the form of a random variable $\alpha_{ij}$ defined below.
\begin{definition}[$\alpha_{ij}$] 
\label{def:alpha}
For every $i\in\LL$ and $j\in\NN_i$, the random variable $\alpha_{ij}\in [0,1]$ represents the probability that 
agent $j$ is a trustworthy neighbor of agent $i$. We assume the availability of such observations $\alpha_{ij}(t)$ at every instant of time $t\geq0$ throughout the paper.
\end{definition}

This model of inter-agent trust observations has been used in prior works~\cite{AURO,ourTRO}. In general, the trust observations can be extracted from the physicality of the systems in various ways as elaborated in the recent survey  \cite{trust_survey_2023}. For example, the works \cite{IJRR,ubicarse} extract trust observation based on the characteristics of the received wireless signals that are used to transfer information between agents. The trust values then capture the likelihood of discrepancies between the agent's reported location and its true location that is manifested in the signal characteristics, such as power and direction of arrival.

The focus of the current work is not on the derivation of the values $\alpha_{ij}$ themselves, but rather on the derivation of a theoretical framework for achieving resilient distributed optimization using this model. Indeed, we show that much stronger results of convergence are achievable by properly exploiting this information in the network. We refer to~\cite{AURO} for an example of such a value $\alpha_{ij}$. Intuitively, a random realization $\alpha_{ij}(t)$ of $\alpha_{ij}\in[0,1]$ 
contains useful trust information regarding the legitimacy of a transmission. 
Here the interval $[0,1]$ denotes the level of trust, where $0$ is the lowest value of trustworthiness and $1$ indicates the highest level of trustworthiness. Thus, a value of $\alpha_{ij}(t)>0.5$ indicates a legitimate transmission, and $\alpha_{ij}(t)<0.5$ indicates a malicious transmission in a stochastic sense (misclassifications are possible).  Note that $\alpha_{ij}(t)=0.5$ means that the observation is completely ambiguous and contains no useful trust information for the transmission at time $t$. We note that the results presented in this paper are not limited to the choice of $0.5$ as the threshold for this ambiguity, and can be generalized to any chosen threshold provided that Assumption \ref{assumption:trust_observation} holds for that threshold.

We use the following assumptions regarding the  underlying distribution trust observations throughout the paper:
\begin{assumption}\label{assumption:trust_observation}
(i) \emph{[First moment homogeneity of trust variables]}
  There are  scalars $E_{\LL}>0$ and $E_{\MM}<0$ such that
    \[E_{\LL}\triangleq\EEop[\alpha_{ij}(t)]-0.5,\qquad
    \hbox{for all $i\in \LL,\ j\in \NN_i\cap\LL$},\] 
    \[E_{\MM}\triangleq\EEop[\alpha_{ij}(t)]-0.5,\qquad
    \hbox{for all $i\in\LL,\ j\in \NN_i\cap\MM$}. \]
(ii) \emph{[Independence of trust observations]}  
    The observations $\alpha_{ij}(t)$ are independent for all $t$ 
    with $i\in \LL$, $j\in\mathcal{N}_i$. 
\end{assumption}
We note that these assumptions are similar to the probabilistic trust framework employed in~\cite{AURO,ourTRO}.

\textit{Discussion:}
For the sake of simplicity of presentation, we assume that the expectation terms $E_{\LL}$ and $E_{\MM}$ are homogeneous (see Assumption  \ref{assumption:trust_observation}(i)). We note that the closed-form analytical results derived in this work can be readily extended to the case where $E_{\LL}\triangleq \min_{i\in \LL, j\in\NN_i\cap\LL}\liminf_{t\rightarrow\infty}\{\EEop[\alpha_{ij}(t)]-0.5\}$ and $E_{\MM}\triangleq \max_{i\in\LL, j\in\NN_i\cap\MM}\limsup_{t\rightarrow\infty}\{\EEop[\alpha_{ij}(t)]-0.5\}$ are such that  $E_{\LL}>0$ and $E_{\MM}<0$. Additionally, our analysis can be readily adapted to incorporate heterogeneity across clients and replace the uniform $E_{\LL}$ and $E_{\MM}$ terms with per link terms, i.e., $E_{ij}=\EEop[\alpha_{ij}(t)]-0.5$.
Additionally, while in this work we derive a first-moment analysis by utilizing Hoeffding's inequality, one can derive an analogous second-moment analysis by utilizing Bernstein's inequality.
Finally, we include in  Section~\ref{sec:learning_trusted_neighbors} a  discussion regarding the relaxation of   Assumption \ref{assumption:trust_observation}(ii)  to include random processes with memory such as Markov chains and martingales.

\subsection{The update rule of agents}

We propose distributed methods with significantly stronger resilience compared to \cite{Sundaram_distributed_opt_malicious_2019}. This is enabled by each legitimate agent learning which neighbors it can trust
while optimizing a system objective.

\paragraph{The update rule for the legitimate agents}
Each legitimate agent $i$ updates $x_i(t)$ by considering the values $x_j(t)$ of its neighbors and {\it the gradient of its own objective function $f_i$} similarly to the iterates described in~\cite{distr-subgrad2009}\footnote{Note, however, that \cite{distr-subgrad2009} does not include the projection on the set $\XX$.}. This method takes the following form for every legitimate agent $i\in\LL$,
    \begin{flalign}\label{eq:legitimate_update_opt_gradient_selfish}
    c_{i}(t)&=w_{ii}(t) x_i(t)+ \sum_{j\in\mathcal{N}_i}w_{ij}(t)x_j(t),\nonumber\\
    y_{i}(t)&=c_i(t)-\gamma(t)\nabla f_i(c_i(t)),\nonumber\\
    x_i(t+1) &= \Pi_{\mathcal{X}}\left(y_i(t)\right),
    \end{flalign}
    where $\gamma(t)\geq 0$ is a stepsize that is common to all agents $i\in\LL$ at each time $t$, and $\Ni$ is the set of neighbors of agent $i$ in the communication graph.
    The set $\Ni$ is composed of both legitimate and malicious neighbors of agent $i\in\LL$, while the weights $w_{ij}(t),\ j\in\Ni\cup\{i\},$ are nonnegative and sum to 1. For each $i\in\LL, j\in\LL\cup\MM$, the choice of $w_{ij}(t)$ depends on the history of the random trust observations $(\alpha_{ij}(\tau))_{0\leq\tau\leq t}$. As a result, the weights $w_{ij}(t)$ and the data points $x_i(t)$ are random, for every $i\in\LL, \, j\in\LL\cup \MM$ and time instant $t>0$.
    
    To increase the convergence rate, we allow $\gamma(t)$ and $w_{ij}(t)$  to depend on an arbitrary parameter $T_0\geq 0$ which dictates how many trust observations a legitimate agent collects before starting to execute the distributed optimization protocol. 
    We specify $\gamma(t)$ and $w_{ij}(t), \:i\in\LL$, $j\in\Ni\cup\{i\}$, precisely later on in Section~\ref{sec:algorithm}. 
\paragraph{The update rule for the malicious agents}
Malicious agents $i\in\MM$ choose values in the set $\XX$ and they may or may not collaborate in choosing their values. We assume that their actions are not known, and thus we do not model them, but assume the worst-case scenario instead in our analysis.  For simplicity of exposition, the dynamic \eqref{eq:legitimate_update_opt_gradient_selfish} captures malicious inputs, where an adversarial agent $i\in\MM$ sends all its legitimate neighbors identical copies of its chosen input $x_i(t)$ at time $t$. Let us denote by $x_{ij}(t)$  the input of a malicious agent $i$ to a legitimate agent $j$ at time $t$, then  $x_{ij_1}(t)=x_{ij_2}(t)$ for every $j_1,j_2\in \Ni\cap\LL$.   
Nonetheless, our analytical results also hold for byzantine inputs where an adversarial agent $i\in\MM$ can send its legitimate neighbors different inputs at time $t$. In this case $x_{ij_1}(t)$ need not be equal to $x_{ij_2}(t)$ for every $j_1,j_2\in \Ni\cap\LL$.

\subsection{Assumptions on connectivity and objective functions}

\begin{assumption}[Sufficiently connected graph]
\label{assumption:connectivity_legitimate}
The set of legitimate agents is not empty, i.e., $\LL\neq\emptyset$. Furthermore, the subgraph $\G_{\LL}$ induced by the legitimate agents is connected.
\end{assumption}

\begin{assumption}\label{assumption:domain_sets}
We assume that $\XX\subset \mathbb{R}^d$ is  compact and convex and that there exists a \textit{known} value $\eta>0$ such that 
\begin{flalign}\label{eq:max_val_norm}
\|x\|\leq \eta,\quad \forall x\in\XX.
\end{flalign}
\end{assumption}
The $\eta$ value in Assumption \ref{assumption:domain_sets} is arbitrary, and its role is to bound the malicious agents' inputs away from infinity.

\begin{assumption}\label{assumption:continuous_gradients_strongly_convex}
For all legitimate agents $i\in\LL$,
the function $f_i$ is $\mu$-strongly convex and 
has $L$-Lipschitz continuous gradients on $\XX$, i.e.,
$\|\nabla f_{i}(x)-\nabla f_{i}(y)\|\leq L\|x-y\| \text{ for all } x,y\in\XX$.
\end{assumption}

Note that under Assumption \ref{assumption:domain_sets} and the strong convexity of Assumption \ref{assumption:continuous_gradients_strongly_convex}, the problem~\eqref{eq:dist_opt_obj} has a unique solution $x^*_{\LL}\in \XX$.

\begin{assumption}\label{assumption:bounded_step_size_series}
Let the stepsize sequence $\{\gamma(t)\}$ be nonnegative, monotonically nonincreasing, and such that
$\sum_{t=0}^{\infty} \gamma(t)=\infty$  and 
$\sum_{t=0}^{\infty} \gamma^2(t)<\infty$.
\end{assumption}

\subsection{Research objectives}
The objective of this work is to arrive at strong convergence results for the distributed optimization problem in~\eqref{eq:dist_opt_obj} in the presence of malicious agents $\MM$. We wish to achieve this by exploiting the availability of stochastic trust values $\alpha_{ij}(t)$ in the network. Specifically, we aim to achieve the following:
\\
\noindent
\ul{\textbf{Objective 1}}: 
We wish to construct weight sequences $\{w_{ij}(t)\}$, $i\in\LL,\ j\in \NN_i$ in the method~\eqref{eq:legitimate_update_opt_gradient_selfish} to weight the influence of neighboring nodes in each legitimate agent's update. Specifically, we want to construct these sequences such that they converge over time to some \emph{nominal weights} $\overline{w}_{ij}, i\in\LL,\ j\in \NN_i,$ almost surely (a.s.), where $\overline{w}_{ij}=0$ 
for all malicious neighbors $j\in\NN_i\cap\MM$ of agent $i\in\LL$.

\noindent
\ul{\textbf{Objective 2}}: Utilizing the proposed weights $\{w_{ij}(t)\}$, we aim to show that the iterates given by \eqref{eq:legitimate_update_opt_gradient_selfish} converge (in some sense) to the true optimal point $x_{\LL}^{\star}\in\XX$ under Assumptions \ref{assumption:trust_observation}-\ref{assumption:bounded_step_size_series}.

\noindent
\ul{\textbf{Objective 3}}: We aim to establish an upper bound on the expected value of $\|x_i(t)-x_{\LL}^{\star}\|^2$, for all $i\in\LL$, as a function of the time $t$, for the iterates $x_i(t)$ produced by the method. 

\section{Algorithm}\label{sec:algorithm}

Next, we present an algorithm 
that incorporates the legitimate agents' learning of inter-agent trust values into the dynamic \eqref{eq:legitimate_update_opt_gradient_selfish} through the choice of the time-dependent weights $w_{ij}(t)$. We recall that these weights depend on the parameter $T_0$.  We utilize the parameter $T_0$ to enable faster convergence rates of the algorithm, nonetheless, $T_0$ is mostly used for analysis and it is not strictly necessary. In other words,
%Algorithm \ref{alg:agent_i_dynamic}.
as we show  in Section~\ref{sec:analytical_results_correct_class_time}, \emph{the algorithm  converges to the optimal point $x_{\LL}^{\star}$ for any choice of nonnegative integer $T_0$, including the special case where $T_0=0$}. In this case,  legitimate agents have no prior trust observations to rely on when they first decide whether to trust their neighbors.

\subsection{The weight matrix sequence}

Consider the sum over a history of $\alpha_{ij}(t)$ values that we denote by $\beta_{ij}(t)$:
\begin{align}
    \label{eq:betas}
    \beta_{ij}(t)=\sum_{k=0}^{t-1} \left(\alpha_{ij}(k)-0.5\right)
\hbox{ \ for $t\ge1,\:i\in\LL,\ j\in\mathcal{N}_i$},
\end{align}
and define $\beta_{ij}(0)=0$. We explore the probabilistic characteristics of $\beta_{ij}(t)$ in Section~\ref{sec:learning_trusted_neighbors}.

We define a time dependent \emph{trusted neighborhood} for agent $i\in\LL$ as:
\begin{align}%\label{def:N_i_t}
\label{eq:Ni_t}
    \NN_i(t) \triangleq \{j\in\NN_i:\:\beta_{ij}(t)\geq0\}.
\end{align} 
This set is the subset of neighbors that legitimate agent $i$ classifies as its legitimate neighbors at time $t$.
For all $t\ge0$, let
\[d_i(t) \triangleq |\NN_i(t)|+1\geq 1\qquad\hbox{for all }i\in\LL.\]
At each time $t$, {\it every agent $i$ sends the value $d_i(t)$
to its  neighbors $j\in \NN_i$ 
in addition to the value $x_i(t)$}. Alternatively, we can assume that agent $i$ sends $d_i(t)$ to its neighbors only when the value $d_i(t)$ changes.
 
Legitimate agents are the most susceptible to making classification errors regarding the trustworthiness of their neighbors when they have a small sample size of trust value observations. Thus, we delay the updating of legitimate agents' values until time $T_0\ge 0$. Up to time $T_0$, the legitimate agents only collect observations of trust values.
 
We define the weight matrix $W(t)$ by choosing its 
entries $w_{ij}(t)$ as follows: for every $i\in\LL$, $j\in\LL\cup\MM$
\begin{align}
\label{eq:weights}
    w_{ij}(t) %\nonumber\\
    = 
    \begin{cases}
    \frac{\mathds{1}_{\{t\geq T_0\}}}{2\cdot\max\{d_i(t),d_j(t)\}} & \text{if } j\in \NN_i(t),\\
    0 & \text{if } j\notin \NN_i(t)\cup \{i\},\\
    1-\displaystyle{\sum_{m\in\NN_i(t)}w_{im}(t)} & \text{if } j=i.
    \end{cases}
\end{align}
Using the weights~\eqref{eq:weights} and letting the stepsize $\gamma(k)=0,\:\forall k < 0$, the dynamic in~\eqref{eq:legitimate_update_opt_gradient_selfish} is equivalent to the following dynamic where  agents \emph{only consider the data values received from their trusted neighbors at time $t$, i.e., $\Ni(t)$,} when computing their own value updates: 
for all $i\in\LL$ and all $t\ge0$,
    \begin{flalign}\label{eq:legitimate_update_opt_gradient_selfish_trusted}
    c_{i}(t)&=w_{ii}(t) x_i(t)+\hspace{-0.45cm} \sum_{j\in\NN_i(t)\cap\LL}\hspace{-0.45cm}w_{ij}(t)x_j(t)+\hspace{-0.45cm}\sum_{j\in\NN_i(t)\cap\MM}\hspace{-0.45cm}w_{ij}(t)x_j(t),\nonumber\\
    y_{i}(t)&=c_i(t)-\gamma(t-T_0)\nabla f_i(c_i(t)),\nonumber\\
    x_i(t+1) &= \Pi_{\XX}\left(y_i(t)\right).
    \end{flalign}

We note that though the choice of the parameter $T_0$ affects the weights $w_{ij}(t),\:i\in\LL,j\in\LL\cup\MM$ and the terms $c_{i}(t),y_{i}(t)$ and $x_i(t), \:i\in\LL$, we omit this dependence from these notations for the sake of clarity of exposition.
\begin{algorithm}[t!]
\caption{The protocol of each agent $i\in\LL$.}\label{alg:agent_i_dynamic}
\begin{algorithmic}
\State \textbf{Inputs:} $T$, $T_0$, $\NN_i$, $x_i(0)$, $\nabla f_i(\cdot)$, $\gamma(\cdot)$.
\State \textbf{Outputs:} $x_i(T)$.
\State Set $\beta_{ij}(t)=0$ for all $j\in\NN_i$;
\For{$t=0,\ldots,T-1$}
\State Set $\NN_i(t) = \{j\in\NN_i:\:\beta_{ij}(t)\geq0\}$;
\State Set $d_i(t) = |\NN_i(t)|+1$;
\State Send $x_i(t)$ and $d_i(t)$ to neighbors;
\For{$j\in\NN_i$}
\State Receive $x_j(t)$ and $d_j(t)$;
\State Extract $\alpha_{ij}(t)$;
\State Set  $\beta_{ij}(t+1)=\beta_{ij}(t)+ \left(\alpha_{ij}(t)-0.5\right)$;
\vspace{0.05cm}
\State Set  the weight $w_{ij}(t)$ based on the values of $T_0$,\phantom . \phantom .\phantom . \phantom .\phantom . \phantom .\phantom . \phantom . \phantom . $\NN_i(t)$, $d_i(t)$, and $d_j(t)$ as follows: 
\[w_{ij}(t)=\frac{\mathds{1}_{\{t\geq T_0\}}\mathds{1}_{\{j\in \NN_i(t)\}}}{2\max\{d_i(t),d_j(t)\}};\] 
\EndFor
\State Set $w_{ii}(t)=1-\sum_{m\in\NN_i}w_{im}(t)$;
\State Set $x_i(t+1)$ according to the dynamic \eqref{eq:legitimate_update_opt_gradient_selfish_trusted};
\EndFor
\end{algorithmic}
\end{algorithm}
 \setlength{\textfloatsep}{0.1cm}

The dependence of the weights $w_{ij}(t)$ on the trust observation history $\beta_{ij}(t)$ comes in through the choice of time-dependent and random trusted neighborhood $\NN_i(t)$ (see~\eqref{eq:Ni_t}). 
 Consequently, some entries of the matrix $W(t)$ are also random,
as seen from~\eqref{eq:weights}. 
The gradients  $\nabla f_i(c_i(t))$ are stochastic due to the randomness of $c_i(t)$, however, they are not unbiased as typically assumed in stochastic approximation methods, including \cite{Stochastic_Subgradient2017}. Thus, we cannot  readily rely on prior analysis for stochastic approximation methods. However, as we show in our subsequent analysis, the variance of 
$\|\nabla f_i(c_i(t))\|$  decays sufficiently fast and allows convergence to the optimal point even in the presence of malicious agents.      
%----------------------------------------------------------
\vspace{-0.1cm}
\section{Learning the sets of trusted neighbors}\label{sec:learning_trusted_neighbors}
This section establishes key characteristics of the stochastic observations $\alpha_{ij}(t)$ that result from the model described in Section~\ref{sec:trustVals} and that we will subsequently use in our analysis of the convergence of the iterates produced by Algo.~\ref{alg:agent_i_dynamic}.  

Recall that we consider the sum $\beta_{ij}(t)$, defined in \eqref{eq:betas}, over a history of $\alpha_{ij}(t)$ values.
Intuitively, following the discussion on $\alpha_{ij}$'s immediately after Definition~\ref{def:alpha}, the values $\beta_{ij}(t)$ will tend towards positive values for legitimate agent transmissions $i\in\LL$ and $\,j\in \NN_i\cap\LL$, and will tend towards negative values for malicious agent transmissions where $i\in\LL$ and $j\in\NN_i\cap\MM$. We restate an important result shown in~\cite{ourTRO} regarding the exponential decay rate of misclassifications given a sum over the history of stochastic observation values that we will use extensively in the forthcoming analysis.

\begin{lemma}[Lemma 2 \cite{ourTRO}]
\label{Lemma:concentration_upper}
Consider the random variables $\beta_{ij}(t)$ capturing the history of stochastic trust values
as defined in~\eqref{eq:betas} and let Assumption \ref{assumption:trust_observation} hold.
Then, for every $t\geq 0$ and every $i\in \LL$, $j\in\NN_i\cap\LL,$
\begin{align*}
\Pr\left(\beta_{ij}(t)< 0\right)\leq \max\{\exp(-2tE_{\LL}^2),\mathds{1}_{\{E_{\LL}<0\}}\},
\end{align*}
while for every $t\geq 0$ and every 
$i\in \LL,\ j\in \NN_i\cap\MM$,
\begin{align*}
\Pr\left(\beta_{ij}(t)\geq 0\right)\leq \max\{\exp(-2tE_{\MM}^2),\mathds{1}_{\{E_{\MM}>0\}}\}.
\end{align*}
\end{lemma}
In other words, the probability of misclassifying malicious agents as legitimate, or vice versa, decays exponentially in the accrued number $t$ of observations.

We can now conclude that there is a random but finite time $T_f$ such that there exists a legitimate agent $i$ which misclassifies the trustworthiness of at least one of its neighbors at time $T_f-1$ whenever $T_f\geq1$, and all the legitimate agents classify the trustworthiness of their neighbors correctly at each time $t$ such that $t\geq T_f\geq0$. We refer to the time $T_f$ as the ``correct classification time". 

Denote by $\mathcal{F}_k$ the filtration of the random trust observation sequences $\{\alpha_{ij}(0),\ldots,\alpha_{ij}(k)\}_{i\in\LL,j\in\LL\cup\MM}$, $k\geq0$.  Interestingly, we observe that $\{T_f=k\}$ cannot be a stopping time\footnote{For the definition of a stopping time see \cite{DU04}.} with respect to $\mathcal{F}_k$. This follows since $\{T_f=k\}$ depends on future trust observations, i.e. after time $k$, and thus $\{T_f=k\}$ is not measurable with respect to the $\sigma$-algebra $\mathcal{F}_k$. Consequently, at each time $t$ we cannot conclude if the classification of even a single link, e.g.,  between $i$ and $j$, is correct.

\begin{corollary}\label{cor:rand_finite_classifications}
Under Assumption \ref{assumption:trust_observation}, there exists a random finite time $T_f\geq0$, almost surely, such that 
\begin{flalign}
\beta_{ij}(t)&\geq0 \text{ for all $t\geq T_f$ and all } i\in\LL,j\in\mathcal{N}_i\cap\LL, \nonumber\\
\beta_{ij}(t)&<0 \text{ for all $t\geq T_f$ and all } i\in\LL,j\in\mathcal{N}_i\cap\MM.
\end{flalign}
Furthermore, if $T_f\geq1$ there exists $i\in\LL$ such that
\begin{flalign}
\beta_{ij}(T_f-1)&<0 \text{ for some } j\in\mathcal{N}_i\cap\LL,\: \text{ or},\nonumber\\ 
\beta_{ij}(T_f-1)&\geq 0 \text{ for some } j\in\mathcal{N}_i\cap\MM.
\end{flalign}
\end{corollary}
\begin{proof}
The proof of this corollary relies on the Borel-Cantelli lemma and follows directly from \cite[Proposition 1]{ourTRO}.
\end{proof}

\textit{Discussion:} The proof of Corollary \ref{cor:rand_finite_classifications} relies on the part of the Borel-Centelli lemma that does not require statistical independence. Thus, a similar conclusion holds for any sequence of probabilities $\{\Pr\left(\beta_{ij}(t)< 0\right)\}_{t=0}^{\infty}$, $i\in\LL,j\in\mathcal{N}_i\cap\LL$ such that $\sum_{t=0}^{\infty}\Pr\left(\beta_{ij}(t)< 0\right)<\infty$. This, for example, holds for dependent trust observations such that $\{\beta_{ij}(t)\}_{t=0}^\infty$ is a submartingle under some additional assumptions. In this case, using the Azuma-Hoeffding inequality (see \cite{Yemini2021_Azuma} for similar calculations for Markov chains) leads to the conclusion that Corollary~\ref{cor:rand_finite_classifications} establishes. A similar discussion is true for classifying malicious neighbors with some adaptations.

For a legitimate agent $i\in\LL$, let $|\NN_i\cap\LL|$ be the number of legitimate neighbors and $|\NN_i\cap\MM|$ be the number of malicious neighbors of agent $i$.
We let $D_{\LL}$ be the total number of legitimate neighbors and let $D_{\MM}$ be the total number of malicious neighbors in the system, with respect to the legitimate agents. That is,
\[D_{\LL}\triangleq\sum_{i\in\LL}|\NN_i\cap\LL|\quad \text{and}
\quad D_{\MM}\triangleq\sum_{i\in\LL}|\NN_i\cap\MM|.\]
Additionally, we define 
the following upper bound on the probability that at least one legitimate agent misclassifies one of its legitimate neighbors as malicious or one of its malicious neighbors as legitimate, when observing $k$ trust values for each of its neighbors
\begin{flalign*}
p_c(k)&\triangleq\mathds{1}_{\{k\geq-1\}}\Bigg[D_{\LL}e^{-2\max\{k,0\}E_{\LL}^2}+D_{\MM}e^{-2\max\{k,0\}E_{\MM}^2}\Bigg].
\end{flalign*}
We define  $p_c(-1)$ in the above equation for the sake of notational completeness of Lemma \ref{lemma:classification_prob}.
Furthermore, we define the following upper bound on the probability that a  legitimate agent misclassifies one of its legitimate or malicious neighbors, in one of the times after observing $k$ trust values for each of its neighbors:
\begin{flalign}\label{eq:def_p_e}
p_e(k)&\triangleq D_{\LL}\frac{\exp(-2kE_{\LL}^2)}{1-\exp(-2E_{\LL}^2)}+D_{\MM}\frac{\exp(-2kE_{\MM}^2)}{1-\exp(-2E_{\MM}^2)}.
\end{flalign}
Using these quantities, we obtain some useful bounds on the probabilities of the events $(T_f=k)$ and $(T_f>k-1)$ for any $k\geq0$, as follows.
\begin{lemma}\label{lemma:classification_prob}
For every $k\geq 0$,  
\begin{flalign}
\Pr(T_f=k) &\leq \min\{p_c(k-1),1\},  \text{ and,} \label{eq:prob_miscalssifying_T_f}\\
\Pr(T_f>k-1) &\leq \min\{p_e(k-1),1\}.\label{eq:T_f_greater_t}
\end{flalign}
\end{lemma}
We present the proof of Lemma~\ref{lemma:classification_prob} in Appendix~\ref{append:proof_lemma_classification_prob}.
Note that \eqref{eq:prob_miscalssifying_T_f} and \eqref{eq:T_f_greater_t} are well defined for $k<0$, since  $p_c(k)\geq0$ and $p_{e}(k)>1$ for all  $k<0$.

\section{Asymptotic Convergence to Optimal Point}\label{sec:analytical_results_correct_class_time}

This section analyzes the convergence characteristics of Algo.~\ref{alg:agent_i_dynamic} by utilizing the almost surely finite correct classification time $T_f$ and the upper bounds of Lemma~\ref{lemma:classification_prob}. 
We show convergence to the nominal optimal point for legitimate agents almost surely and in mean for any finite set of malicious agents and nonempty finite set of legitimate agents.

Assumptions~\ref{assumption:domain_sets} 
and~\ref{assumption:continuous_gradients_strongly_convex} lead to the following conclusion.
\begin{corollary}\label{corollary:bounded_gradients}
When $\XX$ is compact, Assumption \ref{assumption:continuous_gradients_strongly_convex} implies that there is a scalar $G$ such that $\|\nabla f_i(x)\|\leq G,\  \forall x\in\XX,\ i\in \LL$. 
\end{corollary}

The following lemma is a direct consequence of \cite[Lemma 8]{nedicConstrainedConsensus}. Nonetheless, for completeness of presentation we provide the proof.
    \begin{lemma}\label{lemma:upper_bound_phi}
   Denote $\phi_{i}(t)\triangleq\Pi_{\XX}\left(y_i(t)\right)-c_i(t)$. For every $i\in\LL$, $T_0\geq 0$, and $t\geq T_0$ we have that 
    \[\|\phi_{i}(t)\|\leq \gamma(t-T_0)G,\]
    where $G>0$ is from Corollary~\ref{corollary:bounded_gradients}.
%    \sg{I don't recall what $\phi_i$ is.}
    \end{lemma}
    \begin{proof}
    Since $x_i(t)\in\XX$ for all $i\in\LL\cup\MM$ 
    and $t\geq 0$, and $[w_{ij}(t)]_{i\in\LL,j\in\LL\cup\MM}$ is a row stochastic matrix,
    by the convexity of the set $\XX$, we have that  $c_i(t)\in\XX$ for all $i\in\LL$ and $t\geq 0$. Now, 
    %since $\Pi_{\XX}(x)=x$ for every $x\in\XX$, 
    by the standard non-expansiveness property of the projection operator it follows that
    \begin{flalign*}
    \|\phi_{i}(t)\|&=\|\Pi_{\XX}\left(y_i(t)\right)-\Pi_{\XX}\left(c_{i}(t)\right)\|\leq \|y_i(t)-c_{i}(t)\|\nonumber\\
    &=\|c_i(t)-\gamma(t-T_0)\nabla f_i(c_i(t))-c_{i}(t)\|\nonumber\\
    &\leq \gamma(t-T_0)\|\nabla f_i(c_i(t))\| \leq \gamma(t-T_0) G. \hspace{1.5cm}\qedhere
    \end{flalign*}
    \end{proof}

Let us denote $d_{i,\LL}\triangleq|\NN_i\cap\LL|+1$.
Next, we define the doubly stochastic matrix $\overline{W}_{\LL}\in[0,1]^{|\LL|\times|\LL|}$ with the entries $[\overline{W}_{\LL}]_{i,j}$, for every $i,j\in\LL$,
\begin{flalign}\label{eq:weight_no_malicious}
[\overline{W}_{\LL}]_{i,j}= \begin{cases}
    \frac{1}{2\cdot\max\{d_{i,\LL},d_{j,\LL}\}} & \text{if\,}  j\in \NN_i,\\
    0 & \text{if\,}  j\notin \NN_i\cup \{i\},\\
    1-\displaystyle{\sum_{m\in\NN_i\cap\LL}\overline{w}_{im}} & \text{if\,} j=i.
    \end{cases}
\end{flalign}
Note that $\overline{W}_{\LL}$ is the \emph{nominal} weight matrix, i.e., the weight matrix that would be used in the absence of malicious agents. Let $\rho_{\LL}=\sigma_2(\overline{W}_{\LL})$ be the second largest singular value of $\overline{W}_{\LL}$.
Since $\mathbb{G}_{\LL}$ is connected and $\overline{W}_{\LL}$ is symmetric and doubly stochastic,  the value $\rho_{\LL}<1$ is equal to the second largest eigenvalue modulus of $\overline{W}_{\LL}$. 
Additionally, our analysis holds for the lazy Metropolis weight as well, for which by \cite[Lemma 2.2]{Olshevsky2017_lazy_metropolis}, the value $\rho_{\LL}$ can be upper bounded by $(1-1/(71|\LL|^2))$, while \cite{Charron_Bost2020_lazy_metropolis} improves the constant $1/71$ to $1/8$.

Next, define for $t\ge 0$ the following deterministic dynamic that excludes malicious agents:
\begin{flalign}\label{eq:no_malicious_update_opt_gradient_selfish_analysis_simple}
    r_{i}(t)&=\overline{w}_{ii} z_i(t)+ \sum_{j\in\NN_i\cap\LL}\overline{w}_{ij}z_j(t),\nonumber\\
    z_i(t+1) &= \Pi_{\XX}\left(r_i(t)-\gamma(t)\nabla f_i(r_i(t))\right).
    \end{flalign}
    Let $\epsilon\in(0,1)$, $\mu_{\epsilon}\triangleq\mu(1-\epsilon)$, and denote for $T\ge 1$, the auxiliary error function which we utilize to characterize the convergence rate:
\begin{flalign}\label{eq:h_overline_T_def}
&\overline{h}_{\epsilon}(T)\triangleq
\frac{G^2T}{\mu_{\epsilon}}+\frac{4G^2T}{\mu_{\epsilon}(1-\rho_{\LL})}+\frac{2\eta G}{(1-\rho_{\LL})^2}+\frac{4G^2}{\mu_{\epsilon}(1-\rho_{\LL})^3}\nonumber\\
&+2(\mu_{\epsilon}/\epsilon+L)\bigg[\frac{\eta(\eta\mu_{\epsilon}+4G)}{\mu_{\epsilon}(1-\rho_{\LL})^2}+\frac{4G^2\ln\left(\frac{T+2}{2}\right)}{\mu_{\epsilon}^2(1-\rho_{\LL})^2}\nonumber\\
&\hspace{1cm}+\frac{4G(\eta\mu_{\epsilon}+G)}{\mu_{\epsilon}^2(1-\rho_{\LL})^3}+\frac{G^2}{\mu_{\epsilon}^2(1-\rho_{\LL})^4}\bigg].
 \end{flalign}
The function $\bar h(T)$ grows linearly in $T$.  Additionally, it comprises two terms: 1) the first term which captures the contribution from the centralized gradient descent optimization (see  \cite{O_1_t_centralize_strong_convex}) at each time instant without malicious agents (up to a scaling of $1/(1-\epsilon)$), and 2) the following terms that include $\rho_\LL$ which capture the contribution from distributing the optimization over a decentralized network (without malicious agents) that is characterized by the second largest eigenvalue modulus of $\overline{W}_\LL$.
\begin{theorem}\label{theorem:convergence_no_malicious}
If $\LL=\mathbb{V}$ and  Assumptions~\ref{assumption:connectivity_legitimate}-\ref{assumption:bounded_step_size_series} hold, then
 the  dynamic \eqref{eq:no_malicious_update_opt_gradient_selfish_analysis_simple} converges to the optimal point for every initial point $z_i(0)\in\XX$, $i\in\LL$, that is, 
\begin{flalign*}
\lim_{t\rightarrow\infty}\|z_i(t)-x_{\LL}^{\star}\|=0\quad \hbox{for all } i\in\LL.
\end{flalign*}

Moreover, if  $\gamma(t)=\frac{2}{\mu_{\epsilon}(t+2)}$ where $\mu_{\epsilon}\triangleq \mu(1-\epsilon)$, then
\begin{flalign}\label{eq:upper_bound_deviation_no_malicious}
\frac{1}{|\LL|}\sum_{i\in\LL}\|z_i(T)-x_{\LL}^{\star}\|^2\leq  \min\left\{4\eta^2,\frac{4\overline{h}_{\epsilon}(T)}{\mu_{\epsilon} T(T+1)}\right\},
\end{flalign}
for any initial points $z_i(0)\in\XX$, $i\in \LL$,  any $T\geq1$, and any choice of $\epsilon\in(0,1)$.
\end{theorem}
The proof of this theorem is straightforward. Nonetheless, for the sake of the completeness of the presentation, 
we prove Theorem~\ref{theorem:convergence_no_malicious} in 
the supplementary material.

\textit{Discussion:} It may be possible to generalize the explicit upper bound \eqref{eq:upper_bound_deviation_no_malicious} which leads to a convergence rate of $O(1/T)$ to a set of choices for $\gamma(t)$ by extending \cite[Lemma 6]{li2022multi}. Additionally, the work \cite[Corollary 14]{LIU2017162} and the recent work~\cite[Theorem 2.12]{choi2023convergence} relax the individual strong convexity (see Assumption \ref{assumption:continuous_gradients_strongly_convex}) and replace it with the strong convexity of the global objective function at the cost of increasing the convergence rate to $O(1/\sqrt{T})$. Finally, the recent work \cite[
Theorem 2.5]{choi2023convergence} presents conditions to relax Assumption \ref{assumption:domain_sets} (with convergence rate of $O(1/\sqrt{T})$), however, we note that we utilize Assumption \ref{assumption:domain_sets} to limit the inputs of malicious agents.

\subsection{Convergence to optimal point almost surely}

Let us consider the data values $(x_i(t))_{i\in\LL}$ of the dynamic \eqref{eq:legitimate_update_opt_gradient_selfish_trusted}, assuming that $T_f$ is finite. 
Subsequently, for every $t\geq\max\{T_f,T_0\}$ all the legitimate agents participate in the dynamic \eqref{eq:legitimate_update_opt_gradient_selfish_trusted} and all the malicious agents are excluded from it. Thus, from time $\max\{T_f,T_0\}$ the dynamic \eqref{eq:legitimate_update_opt_gradient_selfish_trusted} can be captured by a dynamic of the form \eqref{eq:no_malicious_update_opt_gradient_selfish_analysis_simple}, where $t$ is replaced with $t-\max\{T_f,T_0\}$.

\begin{theorem}[Convergence a.s. to the optimal point]\label{theorem:convergence_a.s.}
If  Assumptions~\ref{assumption:trust_observation}-\ref{assumption:bounded_step_size_series} hold, then the sequence
$\{x_{i}(t)\}$ converges a.s. to $x_{\LL}^{\star}$ for every $i\in\LL$ and $T_0\geq 0$.
\end{theorem}
\begin{proof}
By Corollary \ref{cor:rand_finite_classifications} there exists a finite time $T_f$ such that every legitimate agent $i$ classifies correctly all of its legitimate and malicious neighbors at all times $t\geq T_f$ a.s.
Thus, during this time interval, the dynamic \eqref{eq:legitimate_update_opt_gradient_selfish_trusted} 
%including malicious agents 
is equivalent to the nominal dynamic \eqref{eq:no_malicious_update_opt_gradient_selfish_analysis_simple}
with the initial inputs $z_i(0)=x_{i}(\max\{T_f,T_0\})$ where $i\in\LL$. 
By Theorem~\ref{theorem:convergence_no_malicious}, the dynamic
\eqref{eq:no_malicious_update_opt_gradient_selfish_analysis_simple}  converges to $x_{\LL}^{\star}$.
Additionally, by Assumption  \ref{assumption:domain_sets}, $x_{\LL}(T_f)$  is finite for every finite $T_f$. Applying Corollary \ref{cor:rand_finite_classifications}  with the a.s.~finiteness of $T_f$ concludes the proof.
\end{proof}

\subsection{Convergence to optimal point in mean}
Next, we establish the convergence in mean of each sequence $x_{i}(t)$ to $x_{\LL}^{\star}$, where $i\in\LL$.

\begin{theorem}[Convergence in mean to the optimal point]\label{theorem:convergence_mean}
If  Assumptions~\ref{assumption:trust_observation}-\ref{assumption:bounded_step_size_series} hold, then for every $T_0\geq0$, the sequence $\{x_{i}(t)\}$ converges in the $r$-th mean to $x_{\LL}^{\star}$ for every $i\in\LL$ and $r\geq1$, i.e., 
\[\lim_{t\rightarrow\infty}\EEop\left[\|x_i(t)-x_{\LL}^{\star}\|^r\right]=0\quad \text{ for all } r\geq1.\]
\end{theorem}
We present two types of proofs for this theorem, the first relies on the a.s. convergence of Theorem \ref{theorem:convergence_a.s.} and Assumption  \ref{assumption:domain_sets}. For the sake of completeness of presentation, we additionally establish convergence in mean by definition in  Appendix \ref{proof:converegence_mean_by_definition}. 
\begin{proof}[Proof via Dominated Convergence Theorem]
By the triangle inequality and Assumption~\ref{assumption:domain_sets}, we have $\|x-y\|^{r}\leq (2\eta)^r<\infty$ for all $x,y\in\XX$.
Recalling Theorem~\ref{theorem:convergence_a.s.} we can apply the Dominated  Convergence Theorem (see \cite[Theorem 1.6.7]{DU04}) to each sequence $\{\|x_i(t)-x_{\LL}^{\star}\|^r\}$, $i\in\LL$, to conclude the result. 
\end{proof}

\section{Finite Time Analysis: Expected Convergence Rate}\label{sec:finite_time_analysis}

This section derives analytical guarantees for the finite time regime in the form of the expected convergence rate. We present two upper bounds on the convergence rate. The first upper bound, stated in Theorem \ref{theorem:expected_suboptimality_gap}, relies on  Lemma~\ref{lemma:classification_prob} to provide probabilistic bounds on the correct classification time, and on the convergence rate of the nominal dynamic \eqref{eq:no_malicious_update_opt_gradient_selfish_analysis_simple} which ignores the inputs of malicious agents. We tighten this bound in Theorem \ref{theorem:exp_convergence_rate_alternative} by analyzing the dynamic \eqref{eq:legitimate_update_opt_gradient_selfish_trusted} directly utilizing the bounds on the error probabilities presented in Lemma \ref{Lemma:concentration_upper}.

\subsection{Convergence rate via  correct classification time}

Utilizing Theorem~\ref{theorem:convergence_no_malicious} and Lemma \ref{lemma:classification_prob} we can upper bound the expected suboptimality gap 
by upper bounding the contributions to the suboptimality gap by the event $\{T_f\leq m\}$ and its complement. For each $m$ this leads to an upper bound that coincides with \eqref{eq:upper_bound_deviation_no_malicious} subject to a time shift with $m$ and an additive error term of the probability that $\{T_f> m\}$.
Since for every $m$ either $\{T_f\leq m\}$ is correct or its complement, we can take the minimum over the derived upper bounds.
Before presenting the resulting theorem, recall that $\mu_{\epsilon}\triangleq\mu(1-\epsilon)$ with $\epsilon\in(0,1)$, and that $\eta$ is the maximal input norm, i.e. $\|x_i(t)\|\leq\eta$ for every $i\in\LL\cup\MM$. Additionally, recall that $\overline{h}_{\epsilon}(\cdot)$ is defined in~\eqref{eq:h_overline_T_def} and dictates the convergence rate of the nominal dynamic \eqref{eq:no_malicious_update_opt_gradient_selfish_analysis_simple}. Finally recall the misclassification probability bound $p_e(k)$ defined in \eqref{eq:def_p_e}. 
\begin{theorem}\label{theorem:expected_suboptimality_gap}
Let $\gamma(t)=\frac{2}{\mu_{\epsilon}(t+2)}$. If  Assumptions~\ref{assumption:trust_observation}-\ref{assumption:bounded_step_size_series} hold, then
for every $t\geq T_0$ we have that the average mean squared error is bounded by a decaying function such that for any $\epsilon\in(0,1)$:
\begin{flalign}\label{eq:mal_upper_bound_suboptimality_gap}
&\frac{1}{|\LL|}\sum_{i\in\LL}\EEop[\|x_i(t)-x^{\star}\|^2 ]\leq\nonumber\\ 
&\min_{m\in[T_0:t-1]}\left\{\min\left\{4\eta^2,\frac{4\overline{h}_{\epsilon}\left(t-m\right)}{\mu_{\epsilon}(t-m)(t-m+1)}+4\eta^2 p_e(m)\right\}\right\}.
\end{flalign}
\end{theorem}

Before proving this theorem we present the following corollary by choosing strategic values of $m$ in the objective function of the RHS of \eqref{eq:mal_upper_bound_suboptimality_gap}.

\begin{corollary}\label{corollary:expected_suboptimality_gap}
Let $\gamma(t)=\frac{2}{\mu_{\epsilon}(t+2)}$. For every $t\geq \max\{T_0,1\}+1$ and $\epsilon\in(0,1)$ the mean squared error $\frac{1}{|\LL|}\sum_{i\in\LL}\EEop[\|x_i(t)-x^{\star}\|^2]$ is  upper bounded by $U_1(t,T_0)$ and $U_2(t,T_0)$   
\begin{align}
&U_1(t,T_0)\triangleq \min\left\{4\eta^2,\frac{4\overline{h}_{\epsilon}\left(t-T_0\right)}{\mu_{\epsilon}(t-T_0)(t-T_0+1)}+4\eta^2 p_e\left(T_0\right)\right\},\label{eq:upper_bound_suboptimality_gap_cor1}\\
&\hspace{-0.6cm}U_2(t,T_0) \triangleq\min\left\{4\eta^2,\frac{16\overline{h}_{\epsilon}\left(\frac{t-T_0}{2}\right)}{\mu_{\epsilon}(t-T_0)(t-T_0+2)}+4\eta^2 p_e\left(\frac{t+T_0}{2}-1\right)\right\},\label{eq:upper_bound_suboptimality_gap_cor2}
\end{align}
Further, for $m_3(t)\triangleq \lceil\frac{\ln(t(D_L+D_M))-\ln(1-\exp(-2\min\{E_{\LL}^2,E_{\MM}^2\})}{2\min\{E_{\LL}^2,E_{\MM}^2\}}\rceil$ for every $t$ such that $T_0\leq m_3(t)\leq t-1$, the average mean squared error $\frac{1}{|\LL|}\sum_{i\in\LL}\EEop[\|x_i(t)-x^{\star}\|^2]$ is  upper bounded by 
 \begin{align}
&U_3(t,T_0)\triangleq\min\left\{4\eta^2,\frac{4\overline{h}_{\epsilon}\left(t-m_3(t)\right)}{\mu_{\epsilon}\left(t-m_3(t)\right)\left(t-m_3(t)+1\right)}+ \frac{4\eta^2 }{t}\right\}
,\label{eq:upper_bound_suboptimality_gap_cor3}
\end{align}
for any $\epsilon\in(0,1)$.
\end{corollary}

\begin{proof}
Since $\frac{\overline{h}_{\epsilon}(t)}{t(t+1)}$ and $p_e(t)$ are monotonically decreasing for $t\geq 1$, choosing the values, $m=T_0$, $m=(t+T_0)/2$, and $m=m_3(t)$ in~\eqref{eq:mal_upper_bound_suboptimality_gap}, establishes the proof of Corollary \ref{corollary:expected_suboptimality_gap}.
\end{proof}
We note that since both $\frac{\overline{h}_{\epsilon}(t)}{t(t+1)}$ and $p_e(t)$ are monotonically decreasing with $t$ for all $t\geq 1$, and $\mu_{\epsilon}>0$ the functions $\frac{\overline{h}_{\epsilon}(t-m)}{(t-m)(t-m+1)}$ and $p_e(m)$ have opposite monotonicity with respect to $m$. The three chosen values reflect this observation. The first, i.e., $m=T_0$, is aimed to minimize the term $\frac{\overline{h}_{\epsilon}(t-m)}{(t-m)(t-m+1)}$ at the expense of $p_e(m)$ which for this choice of $m=T_0$ is a constant with respect to $t$. This choice of $m$ is favorable when $T_0$ is close to $t$. The choice $m=m(t)=(t+T_0)/2$ is selected  so that both the terms $\frac{\overline{h}_{\epsilon}(t-m)}{(t-m)(t-m+1)}$ and $p_e(t)$ decrease as a function of $t$, at a linear and an exponential rate, respectively. The last choice, i.e., $m=m_3(t)$, balances the convergence rate of the two additive terms that compose the objective function of \eqref{eq:mal_upper_bound_suboptimality_gap} and thus is best to use for large values of $t$.

Now, we proceed to prove Theorem \ref{theorem:expected_suboptimality_gap} utilizing the law of total expectation with respect to the event $\{T_f\leq m\}$. Additionally, we provide in Section~II of the supplementary material an alternative proof for Theorem \ref{theorem:expected_suboptimality_gap} that considers the individual events $M_k\triangleq\max\{k,T_0\}$.

\begin{proof}[Proof of Theorem \ref{theorem:expected_suboptimality_gap}]
For every $t-1\geq T_0$ and $m\in [T_0:t-1]$ we have that
\begin{flalign*}
&\frac{1}{|\LL|}\sum_{i\in\LL}\EEop\left[\|x_{i}(t)-x_{\LL}^{\star}\|^2\right]\nonumber\\
&\stackrel{(a)}{=} \Pr(T_f\leq m)\frac{1}{|\LL|}\sum_{i\in\LL}\EEop\left[\|x_{i}(t)-x_{\LL}^{\star}\|^2\big|T_f\leq m\right]\nonumber\\
&\quad+\Pr(T_f> m)\frac{1}{|\LL|}\sum_{i\in\LL}\EEop\left[\|x_{i}(t)-x_{\LL}^{\star}\|^2\big|T_f>m\right]\nonumber\\
&\stackrel{(b)}{\leq} \max_{T_f\leq m}\left\{\frac{1}{|\LL|}\sum_{i\in\LL}\EEop\left[\|z_i(t-T_f)-x_{\LL}^{\star}\|^2\:\big|T_f\right]\right\}+4\eta^2\Pr(T_f>m)\nonumber\\
& \stackrel{(c)}{\leq} \min\left\{4\eta^2,\frac{4\overline{h}_{\epsilon}\left(t-m\right)}{\mu_{\epsilon}(t-m)(t-m+1)}\right\}+4\eta^2 p_e(m),
\end{flalign*}
where $(a)$ follows from the law of total expectation, $(b)$ follows from the definition of $T_f$ and Assumption  \ref{assumption:domain_sets}, and $(c)$ follows from the fact that $\frac{\overline{h}_{\epsilon}(t)}{t(t+1)}$ is nonincreasing for $t\geq1$ and Lemma~\ref{lemma:classification_prob}.
\end{proof}

\subsection{Convergence rate via trustworthiness\\ 
misclassification error probabilities}\label{sec:analytical_results_misclassification_error}
This section aims to tighten the bound on the convergence rate presented in Theorem \ref{theorem:expected_suboptimality_gap} and Corollary \ref{corollary:expected_suboptimality_gap}. To this end, we develop an alternative analytical approach that evaluates more carefully how our choice of weights $w_{ij}(t)$, stepsize $\gamma(t)$, and $T_0$, together with the quality of trust values, captured by the constants $E_{\LL}$ and $E_{\MM}$, affects the dynamic \eqref{eq:legitimate_update_opt_gradient_selfish_trusted}.

\begin{lemma}\label{lemma:upper_boud_w_t_diff}
Denote $\overline{c}_i(t) \triangleq \overline{w}_{ii} x_i(t)+ \sum_{j\in\NN_i\cap\LL}\overline{w}_{ij}x_j(t)$. 
Let $r>0$, $i\in\LL$, and $t\geq0$. Then, 
\begin{flalign*}
\EEop\bigg[\|c_i(t)-\overline{c}_i(t)\|^r\bigg]\leq (2\eta)^rp_c(t).
\end{flalign*}
\end{lemma}
\begin{proof}
First, recall that by Assumption \ref{assumption:domain_sets} 
and the stochasticity of the matrices $(w_{ij})_{i\in\LL,j\in\LL\cup\MM}$ and $\overline{W}_{\LL}$ we have that $\|c_i(t)\|\leq \eta$ and $\|\overline{c}_i(t)\|\leq\eta$ for every  $i\in\LL$ and $t\geq0$. Additionally, let $W_{\LL}(t)\triangleq(w_{ij}(t))_{i,j\in\LL}$, then by definition $c_i(t)=\overline{c}_i(t)$ for all $i\in\LL$ whenever $W_{\LL}(t)=\overline{W}_{\LL}$. Consequently, by the law of total expectation:
\begin{flalign}
&\EEop\bigg[\|c_i(t)-\overline{c}_i(t)\|^r\bigg]\nonumber\\
&=
\Pr(W_{\LL}(t)=\overline{W}_{\LL})\EEop\bigg[\|c_i(t)-\overline{c}_i(t)\|^r\big|W_{\LL}(t)=\overline{W}_{\LL}\bigg]\nonumber\\
&\quad +\Pr(W_{\LL}(t)\neq\overline{W}_{\LL})\EEop\bigg[\|c_i(t)-\overline{c}_i(t)\|^r\big|W_{\LL}(t)\neq\overline{W}_{\LL}\bigg]\nonumber\\
&\leq (2\eta)^rp_c(t).\nonumber \hspace{6cm}\qedhere
\end{flalign}
\end{proof} 

\textit{Discussion:}
We used the term $p_c(k)$ in Lemma \ref{lemma:upper_boud_w_t_diff} to avoid introducing an additional notation.
Nevertheless, we can tighten this lemma by observing that in the derivation, we should only consider the neighbors of the legitimate agent $i$. Thus, for the purpose of Lemma \ref{lemma:upper_boud_w_t_diff}, the constant $D_{\LL}$ and $D_{\MM}$ in $p_c(k)$ can be replaced with the notations $d_{\LL}\triangleq \max_{i\in\LL} |\Ni\cap\LL|$ and $d_{\MM}\triangleq \max_{i\in\LL} |\Ni\cap\MM|$, respectively.

Next, we present an auxiliary proposition that we utilize in upper bounding the expected distance between an agent's value and the average agents' values at time $t$. 
Here we extend \cite[Lemma 11]{rabbat-tradeoffs} to the case of $d$-dimensional vectors with \textit{random} perturbations. Note that a na\"ive implementation of \cite[Lemma 11]{rabbat-tradeoffs} for each of the dimensions $1,\dots,d$ scales the resulting upper bound by $\sqrt{d}$. The upper bound we next derive eliminates this scaling.

Let $A\in\mathbb{R}^{d\times|\LL|}$ and denote by $[A]_j$ the $j$th column of $A$.  The  Frobenius norm of the matrix $A$ is defined as
 \[\|A\|_{\text{F}}\triangleq\sqrt{\sum_{i=1}^d\sum_{j\in\LL}|a_{ij}|^2}=\sqrt{\sum_{j\in\LL}\|[A]_j\|^2}=\sqrt{\sum_{i=1}^d\|[A^T]_i\|^2},\]
 where $A^T$ denotes the transpose matrix of the matrix $A$. Additionally, we denote by $\boldsymbol{1}$ the all-ones column vector with $|\LL|$ entries. 
 
\begin{proposition}\label{prop:distance_to_mean_time_t}
Let $\XX\subset \mathbb{R}^d$  fulfill Assumption \ref{assumption:domain_sets}. 
Additionally, let $W(t)\triangleq(w_{ij}(t))_{i,j\in\LL}$ be deterministic doubly stochastic matrices such that $\sigma_2(W(t))\leq \rho$ for all $t\geq 0$. Furthermore, let $\Delta_i(t)\in\mathbb{R}^{d\times 1}$ be random vectors, and let $X(t)\in\mathbb{R}^{d\times |\LL|}$  be defined by the following dynamic 
\begin{flalign}
X(t+1)&= X(t)W^T(t)+\Delta(t),
\end{flalign}
where $\Delta(t)=(\Delta_1(t),\ldots,\Delta_{|\LL|}(t))$ for all $t\ge0$.
Assume that there exists a non-increasing sequence $\delta(t)$ such that 
\[\EEop[\|\Delta_i(t)\|^2]\leq \delta^2(t), \:\:\forall\: i\in\LL,\]
and let  $t/2\triangleq\lfloor\frac{t}{2}\rfloor$ and
\[\overline{X}(t)\triangleq \frac{X(t)\boldsymbol{1}}{|\LL|}\boldsymbol{1}^T =\frac{1}{|\LL|}\sum_{i\in\LL}[X(t)]_i\boldsymbol{1}^T.\]
Then, for every $X(0)$ such that $[X(0)]_i\in\XX,
\:\:\forall\: i\in\LL$:
\begin{flalign*}
\hspace{-0.2cm}\frac{\sum_{j\in\LL}\EEop[\|[X(t)]_j-\overline{X}(t)\|]}{|\LL|}&\leq 2\eta\rho^t+\frac{\delta(0)\rho^{t/2}}{1-\rho}+\frac{\delta(t/2)}{1-\rho},\\
\hspace{-0.2cm}\frac{\sum_{j\in\LL}\EEop[\|[X(t)]_j-\overline{X}(t)\|^2]}{|\LL|}&\leq \left[2\eta\rho^{t}+\frac{\delta(0)\rho^{t/2}}{1-\rho}+\frac{\delta(t/2)}{1-\rho}\right]^2.
\end{flalign*}
\end{proposition}

We present the proof for Proposition \ref{prop:distance_to_mean_time_t} in Appendix \ref{append:ditance_t_mean_proof}.
We apply this proposition in establishing the upper bounds in the forthcoming lemma.
Toward providing the lemma,
define \[\overline{x}_{\LL}(t)\triangleq \frac{1}{|\LL|}\sum_{i\in\LL}x_i(t), \text{ and }\] 
\begin{flalign}
\delta_{\MM}(t,T_0)&\triangleq 2\eta\rho_{\LL}^{t-T_0}+\frac{(2\eta\sqrt{p_c(T_0)}+G\gamma(0))\rho_{\LL}^{(t-T_0)/2}}{1-\rho_{\LL}}\nonumber\\
&\quad+\frac{2(\eta\sqrt{p_c((t+T_0)/2)}+G\gamma((t-T_0)/2))}{1-\rho_{\LL}}.\label{eq:g_tilde_t_T_0_def}
\end{flalign}

\begin{lemma}\label{lemma:upper_bound_exp_dist_to_mean}
For every $t\geq0$
\begin{flalign*}
&\frac{1}{|\LL|}\sum_{i\in\LL}\EEop[\|x_i(t)-\overline{x}_{\LL}(t)\|]\leq \delta_{\MM}(t,T_0),\text{ and }\nonumber\\
&\frac{1}{|\LL|}\sum_{i\in\LL}\EEop[\|x_i(t)-\overline{x}_{\LL}(t)\|^2]\leq \delta_{\MM}^2(t,T_0).
\end{flalign*}
\end{lemma}

\begin{proof}
Recall that $\phi_{i}(t)\triangleq\Pi_{\XX}\left[y_i(t)\right]-c_{i}(t)$.
The matrices $W_{\LL}(t)\triangleq(w_{ij}(t))_{i,j\in\LL}$ are random,  vary with time, and can be sub-stochastic. Thus we cannot na\"ively apply Proposition \ref{prop:distance_to_mean_time_t}  for $\Delta_i(t)= \phi_{i}(t)$. Instead, we substitute
\begin{flalign*}
\Delta_i(t)=c_i(t)-\overline{c}_i(t)+\phi_i(t),
\end{flalign*}
where $\overline{c}_i(t) \triangleq \overline{w}_{ii} x_i(t)+ \sum_{j\in\NN_i\cap\LL}\overline{w}_{ij}x_j(t)$.
It follows that 
\[x_i(t+1)=\overline{w}_{ii} x_i(t)+ \sum_{j\in\NN_i\cap\LL}\overline{w}_{ij}x_j(t)+\Delta_i(t).\]

By the Cauchy–Schwarz inequality for the $\ell_2$ inner product, and by the Cauchy–Schwarz inequality  \textit{for expectations} 
\begin{flalign*}
\EEop\left[\|\Delta_i(t)\|^2\right] &= \EEop\left[\|c_i(t)-\overline{c}_i(t)+\phi(t)\|^2\right]\nonumber\\
&\leq\EEop\left[\|c_i(t)-\overline{c}_i(t)\|^2\right]+\EEop\left[\|\phi_i(t)\|^2\right]\nonumber\\
&\quad +2\sqrt{\EEop\left[\|c_i(t)-\overline{c}_i(t)\|^2\right]}\sqrt{\EEop\left[\|\phi_i(t)\|^2\right]}.
\end{flalign*}
By Lemma \ref{lemma:upper_boud_w_t_diff} and Assumption \ref{assumption:domain_sets}, we further have
\begin{flalign*}
&\EEop\bigg[\|c_i(t)-\overline{c}_i(t)\|^2\bigg]
\leq 4\eta^2p_c(t).
\end{flalign*}
Additionally, by Assumption  \ref{assumption:continuous_gradients_strongly_convex}  
and the non-expansiveness property of the projection, since $c_i(t)\in \XX$ for all $i$ and all $t$, it follows that
$\|\phi_i(t)\|\leq\|y_i(t)-c_i(t)\| \leq G\gamma(t-T_0)$. Hence,
\begin{flalign}
&\EEop\left[\|\Delta_i(t)\|^2\right] = \EEop\left[\|c_i(t)-\overline{c}_i(t)+\phi_i(t)\|^2\right]\nonumber\\
&\leq 4\eta^2p_c(t)+G^2\gamma^2(t-T_0)+4G\eta\sqrt{p_c(t)}\,\gamma(t-T_0)\nonumber\\
&=\left(2\eta\sqrt{p_c(t)}+G\gamma(t-T_0)\right)^2.
\end{flalign}
We conclude the proof by substituting  $\delta(t)=\widetilde{\delta}(t+T_0)=2\eta\sqrt{p_c(t+T_0)}+G\gamma(t)$, $W(t)=\overline{W}_{\LL}$, and $\rho=\rho_{\LL}$ in Proposition~\ref{prop:distance_to_mean_time_t} and using the transformation  $t\rightarrow t-T_0$.
\end{proof}

Now, we are ready to present our tightened convergence rate guarantees for Algo.~\ref{alg:agent_i_dynamic}.
\begin{theorem}\label{theorem:exp_convergence_rate_alternative} %\snote{Do we require Assumptions 1.(i),(ii), and (iii) here?}
If  Assumptions \ref{assumption:trust_observation}-\ref{assumption:bounded_step_size_series} hold, then
Algo.~\ref{alg:agent_i_dynamic} converges to the optimal point $x^*_\LL$ in the mean-squared sense
for every collection $x_i(0)\in\XX$, $i\in\LL$,
of initial points  i.e.,
\begin{flalign}\label{eq:convergence_alternative_square}
\lim_{t\rightarrow\infty}\EEop\left[\|x_i(t)-x_{\LL}^{\star}\|^2\right]=0,\quad\forall i\in\LL.
\end{flalign}
%whenever $\sum_{t=0}^{\infty}\gamma(t)=\infty$ and $\sum_{t=0}^{\infty}\gamma^2(t)<\infty$.

Moreover, let $\gamma(t)=\frac{2}{\mu_{\epsilon}(t+2)}$. Then, for every $T_0\geq 0$ and $\epsilon\in(0,1)$ there exists a function $C_{\MM}(T_0)$ that decreases exponentially with $T_0$ and is independent of $T$ such that for any collection $x_i(0)\in\XX$, $i\in\LL$,  $\epsilon\in(0,1)$ and for all $T\ge T_0$,
\begin{flalign}\label{eq:upper_bound_suboptimality_gap_alternative}
&\frac{1}{|\LL|}\sum_{i\in\LL}\EEop\left[\|x_i(T)-x_{\LL}^{\star}\|^2\right]\leq \min\left\{4\eta^2,  \frac{4\overline{h}_{\epsilon}(T-T_0)+ C_{\MM}(T_0)}{\mu_{\epsilon}(T-T_0)(T-T_0+1)}\right\},
\end{flalign}
where $\overline{h}_{\epsilon}(\cdot)$ is defined in~\eqref{eq:h_overline_T_def} and $\mu_{\epsilon}\triangleq \mu(1-\epsilon)$.
\end{theorem}
For the sake of simplicity of exposition, we only characterize the function $C_{\MM}(T_0)$ with respect to its exponential decrease in $T_0$. Nonetheless, we define the function $C_{\MM}(T_0)$ in \eqref{eq:def_C_MMT_0} as part of the proof of Theorem~\ref{theorem:exp_convergence_rate_alternative}.
Intuitively, the $C_{\MM}(T_0)$ term above represents the error term contributed by the presence of malicious agents in the distributed network, where the starting time $T_0$ is captured by the shift by $T_0$ with respect to \eqref{eq:upper_bound_deviation_no_malicious}. It can be seen that for a sufficiently large  $T$ the entire term on the right of the inequality~\eqref{eq:upper_bound_suboptimality_gap_alternative} decays on the order of $O\left(\frac{1}{T}\right)$.
Finally, we can observe that Theorem~\ref{theorem:exp_convergence_rate_alternative} tightens the bound \eqref{eq:mal_upper_bound_suboptimality_gap} that is derived in Theorem \ref{theorem:expected_suboptimality_gap} for the regime $T\gg 1$. Consequently, it also tightens the bounds \eqref{eq:upper_bound_suboptimality_gap_cor1}-\eqref{eq:upper_bound_suboptimality_gap_cor3} that are presented in Corollary \ref{corollary:expected_suboptimality_gap} in that regime. 

Before proceeding to prove this theorem, we point out that unlike the analysis for stochastic gradient models such as \cite{Stochastic_Subgradient2017}, in our model $w_{ij}(t)$ and $x_j(t)$ are correlated. This follows by the statistical dependence of $w_{ij}(t)$ and $w_{ij}(t-1)$. Thus, we cannot use the standard analysis which requires that  $\EEop[w_{ij}(t)x_j(t)]=\EEop[w_{ij}(t)]\EEop[x_j(t)]$. Finally, we observe that the nonnegativity of the variance of random variables \eqref{eq:convergence_alternative_square} and the sandwich theorem imply that $\lim_{t\rightarrow\infty}\EEop[\|x_i(t)-x_{\LL}^{\star}\|]=0,\:\forall \: i\in\LL$. This result also holds since convergence in expectation in the $r$th moment implies convergence in expectation in the $s$th moment whenever $0<s<r$.
\vspace{-0.1cm}
\begin{proof}
By the non-expansiveness property of  the projection, for every $t\geq T_0$ and $i\in\LL$,
\begin{flalign*}
\|x_{i}(t+1)-x_{\LL}^{\star}\|^2=\|\Pi_{\XX}\left(y_i(t)\right)-x_{\LL}^{\star}\|^2\leq
\|y_i(t)-x_{\LL}^{\star}\|^2.
\end{flalign*}
Denote 
\begin{flalign*}
\overline{c}_i(t) &\triangleq \overline{w}_{ii} x_i(t)+ \sum_{j\in\NN_i\cap\LL}\overline{w}_{ij}x_j(t),\text{ and }\nonumber\\
\overline{g}_i(t)&\triangleq\overline{c}_i(t)-\gamma(t-T_0)\nabla f_i(\overline{c}_i(t))-x_{\LL}^{\star}.
\end{flalign*}
Recall that $y_i(t)=c_i(t)-\gamma(t-T_0)\nabla f_i(c_i(t))$, then by the Cauchy-Schwarz inequality for the inner product on $\ell_2$, the triangle inequality, and the linearity of the expectation
\begin{flalign*}
&\frac{1}{|\LL|}\sum_{i\in\LL}\EEop\left[\|x_i(t+1)-x_{\LL}^{\star}\|^2\right]\nonumber\\
&= \frac{1}{|\LL|}\sum_{i\in\LL}\EEop\bigg[\|\overline{c}_i(t)-\gamma(t-T_0)\nabla f_i(\overline{c}_i(t))-x_{\LL}^{\star}\nonumber\\
&\quad+c_i(t)-\overline{c}_i(t)+\gamma(t-T_0)[\nabla f_i(\overline{c}_i(t))-\nabla f_i(c_i(t))]\|^2\bigg]\nonumber\\
&\leq \underbrace{\frac{1}{|\LL|}\sum_{i\in\LL}\EEop\bigg[\|\overline{g}_i(t)\|^2\bigg]}_{\text{(I)}}+\frac{1}{|\LL|}\sum_{i\in\LL}\underbrace{\EEop\bigg[\|c_i(t)-\overline{c}_i(t)\|^2\bigg]}_{\text{(II)}}\nonumber\\
&+\frac{\gamma^2(t-T_0)}{|\LL|}\sum_{i\in\LL}\underbrace{\EEop\bigg[\|\nabla f_i(\overline{c}_i(t))-\nabla f_i(c_i(t))\|^2\bigg]}_{\text{(III)}}\nonumber\\
&+ \frac{2}{|\LL|}\sum_{i\in\LL}\underbrace{\EEop\bigg[\|\overline{g}_i(t)\|\cdot \|c_i(t)-\overline{c}_i(t)\|\bigg]}_{\text{(IV)}}\nonumber\\
&+ \frac{2\gamma(t-T_0)}{|\LL|}\sum_{i\in\LL}\underbrace{\EEop\bigg[\|\overline{g}_i(t)\|\cdot \|\nabla f_i(\overline{c}_i(t))-\nabla f_i(c_i(t))\|\bigg]}_{\text{(V)}}\nonumber\\
&+\frac{2\gamma(t-T_0)}{|\LL|}\sum_{i\in\LL}\EEop\bigg[\|c_i(t)-\overline{c}_i(t)\|\nonumber\\
&\hspace{2.6cm}\underbrace{\hspace{1cm}\cdot \|\nabla f_i(\overline{c}_i(t))-\nabla f_i(c_i(t))\|\bigg]}_{\text{(VI)}}.
\end{flalign*}
By~\eqref{eq:max_val_norm}, Lemma~\ref{lemma:upper_boud_w_t_diff} 
we have that 
$\text{(II)}
\leq 4\eta^2p_c(t)$.

Since $\nabla f_i$ are $L$-Lipschitz continuous
\begin{flalign*}
\text{(III)}&\leq L^2\EEop\bigg[\|\overline{c}_i(t)-c_i(t)\|^2\bigg]\leq  4L^2\eta^2p_c(t).
\end{flalign*}
Now, by Lemma \ref{lemma:upper_boud_w_t_diff}$,
\EEop\bigg[\|c_i(t)-\overline{c}_i(t)\|\bigg]\leq 2\eta p_c(t)$.
Thus, 
 \begin{flalign*}
\text{(IV)}&\leq (2\eta+\gamma(t-T_0)G)\EEop\bigg[\|\ \overline{c}_i(t)- c_i(t)\|\bigg]\nonumber\\
&\quad\leq 2\eta\left(2\eta+\gamma(t-T_0)G\right) p_c(t),
\end{flalign*}
and by the $L$-Lipschitz continuity of $\nabla f_i$
 \begin{flalign*}
\text{(V)}&\leq 2L\eta\left(2\eta+\gamma(t-T_0)G\right) p_c(t),\nonumber\\
\text{(VI)}&\leq L\EEop\bigg[\|c_i(t)-\overline{c}_i(t)\|^2\bigg]
\leq 4L\eta^2p_c(t).
\end{flalign*}
Define $h_{\MM}(t,T_0)$  as
\begin{flalign*}
&h_{\MM}(t,T_0)\triangleq 4 \eta^2 p_c(t)\left[3+\frac{(G+4L\eta)\gamma(t-T_0)}{\eta}+\frac{L(G+L\eta)\gamma^2(t-T_0)}{\eta}\right],
\end{flalign*}
and recall that $\overline{x}_{\LL}(t)\triangleq \frac{1}{|\LL|}\sum_{i\in\LL}x_i(t)$.
Therefore, let $\epsilon\in(0,1)$
\begin{flalign*}
&\frac{1}{|\LL|}\sum_{i\in\LL}\EEop\left[\|x_i(t+1)-x_{\LL}^{\star}\|^2\right]\nonumber\\
&\leq \text{(I)}+h_{\MM}(t,T_0)\nonumber\\
&\leq h_{\MM}(t,T_0)+\frac{ \left(1-\mu(1-\epsilon)\gamma(t-T_0)\right)}{|\LL|}\sum_{i\in\LL}\EEop\left[\|x_i(t)-x_{\LL}^{\star}\|^2\right]\nonumber\\
&+\gamma^2(t-T_0)G^2+\frac{2\gamma(t-T_0)}{|\LL|}\sum_{i\in\LL}\Bigg[G\EEop\left[\|x_i(t)-\overline{x}_{\LL}(t)\|\right]\nonumber\\
&\hspace{3.5cm}+\frac{\mu(1/\epsilon-1)+L}{2}\EEop\left[\|x_i(t)-\overline{x}_{\LL}(t)\|^2\right]\Bigg],
\end{flalign*}
where the last inequality follows from Assumption \ref{assumption:continuous_gradients_strongly_convex}, the convexity of $\|\cdot\|^2$, the double stochasticity of $\overline{W}_{\LL}$, and Young inequality (for further details see the proof of Theorem \ref{theorem:convergence_no_malicious}). 

Recall \eqref{eq:g_tilde_t_T_0_def} and the notation $\mu_{\epsilon}\triangleq \mu(1-\epsilon)$.
Denote
\[\widetilde{h}_{\MM}(t,T_0)\triangleq  \gamma(t-T_0)G^2+2G\delta_{\MM}(t,T_0)+(\mu_{\epsilon}/\epsilon+L)\delta_{\MM}^2(t,T_0).\]
Here, the term $\widetilde{h}_{\MM}(t,T_0)$ is affected by the distributed nature of our optimization process and the presence of malicious agents.
We  utilize  Lemma \ref{lemma:upper_bound_exp_dist_to_mean} to conclude that
\begin{flalign*}
&\frac{1}{|\LL|}\sum_{i\in\LL}\EEop\left[\|x_i(t+1)-x_{\LL}^{\star}\|^2\right]\leq \gamma(t-T_0)\widetilde{h}_{\MM}(t,T_0)\nonumber\\
&+h_{\MM}(t,T_0)+\frac{ \left(1-\mu_{\epsilon}\gamma(t-T_0)\right)}{|\LL|}\sum_{i\in\LL}\EEop\left[\|x_i(t)-x_{\LL}^{\star}\|^2\right].
\end{flalign*}
Thus, since $|\LL|<\infty$, 
\begin{flalign}
\lim_{t\rightarrow\infty}\EEop\left[\|x_i(t)-x_{\LL}^{\star}\|^2\right]= 0,\:\forall \: i\in\LL,
\end{flalign}
whenever $\sum_{t=0}^{\infty}\gamma(t)=\infty$ and $\sum_{t=0}^{\infty}\gamma^2(t)<\infty$.

To prove the second part of the theorem, motivated by \cite{O_1_t_centralize_strong_convex} we let $\gamma(t)=\frac{2}{\mu_{\epsilon}(t+2)}$. It follows that 
\begin{flalign*}
&\frac{1}{|\LL|}\sum_{i\in\LL}\EEop\left[\|x_i(t+1)-x_{\LL}^{\star}\|^2\right]
\leq \frac{2\widetilde{h}_{\MM}(t,T_0)}{\mu_{\epsilon}(t-T_0+2)}\nonumber\\
&\:\:+h_{\MM}(t,T_0) +\frac{t-T_0}{t-T_0+2}\cdot\frac{ 1}{|\LL|}\sum_{i\in\LL}\EEop\left[\|x_i(t)-x_{\LL}^{\star}\|^2\right].
\end{flalign*}
Multiplying both sides by $(t-T_0+1)(t-T_0+2)$ and summing over the set $t\in\{T_0,T_0+1,\ldots,T-1\}$ yield the upper bound
    \begin{flalign*}
    &\frac{1}{|\LL|}\sum_{i\in\LL}\EEop[\|x_i(T)-x_{\LL}^{\star}\|^2]
\leq\frac{2\sum_{t=T_0}^{T-1}(t-T_0+1)\widetilde{h}_{\MM}(t,T_0)}{\mu_{\epsilon}(T-T_0)(T-T_0+1)}\nonumber\\
&\hspace{2cm}+\frac{\sum_{t=T_0}^{T-1}(t-T_0+1)(t-T_0+2)h_{\MM}(t,T_0)}{(T-T_0)(T-T_0+1)}.
    \end{flalign*}
Using the identity $\sqrt{a+b}\leq \sqrt{a}+\sqrt{b}$, we deduce that
\begin{flalign}
\sqrt{p_c(t)}&\leq\sqrt{ D_{\LL}}e^{-tE_{\LL}^2}+\sqrt{D_{\MM}}e^{-tE_{\MM}^2}.
\end{flalign}
In addition,   we utilize the identities for  $v\in(0,1)$
\[\sum_{t=0}^{\infty}tv^t=\frac{v}{(1-v)^2} \text{ and } \sum_{t=0}^{\infty}t^2v^t=\frac{v(v+1)}{(1-v)^3}.\]
Denote,
\begin{flalign*}
&\widetilde{C}_1(T_0,E,D)\triangleq \frac{16\eta e^{-T_0E^2}\sqrt{D} }{1-\rho_{\LL}}\Bigg[\frac{2\eta(1-\rho_{\LL})+G/\epsilon+L/\mu_{\epsilon}}{(1-\rho_{\LL})(1-\rho_{\LL}e^{-E^2})} +\nonumber\\
&\frac{G+2(\eta+\frac{G}{\mu_{\epsilon}})(\mu_{\epsilon}/\epsilon+L)}{(1-\rho_{\LL})^2}+\frac{(\mu_{\epsilon}/\epsilon+L)(
G/\mu_{\epsilon}+2\eta\sqrt{D}e^{-T_0E^2})}{(1-\rho_{\LL})^3}\nonumber\\
&\frac{G}{(1-e^{-E^2})^2}+\frac{2(\mu_{\epsilon/\epsilon+L})(G(1-e^{-E^2})+\eta\sqrt{D}e^{-T_0E^2})}{(1-\rho_{\LL})(1-e^{-E^2})^2}\Bigg],
\end{flalign*}
and
$C_1(T_0)\triangleq\widetilde{C}_1(T_0,E_{\LL},D_{\LL})+\widetilde{C}_1(T_0,E_{\MM},D_{\MM})$.

Further algebra yields that
\[\sum_{t=T_0}^{T-1}(t-T_0+1)\widetilde{h}_{\MM}(t,T_0)\leq 2\overline{h}_{\epsilon}(t-T_0)+C_1(T_0),\]
here, the added term $C_1(T_0)$ captures the influence of the malicious agents on the term (I).
Additionally, denote
\begin{flalign*}
&\widetilde{C}_2(T_0,E,D) \triangleq
 \frac{8\eta De^{-2T_0E^2}}{\left(1-e^{-2E^2}\right)}\nonumber\\ 
&\:\:\cdot\Bigg[\frac{3\eta}{\left(1-e^{-2E^2}\right)^2}+\frac{G+4L\eta}{\mu_{\epsilon}(1-e^{-2E^2})}+\frac{2L(G+L\eta)}{\mu_{\epsilon}^2}\Bigg],
\end{flalign*}
and
$C_2(T_0)\triangleq\widetilde{C}_2(T_0,E_{\LL},D_{\LL})+\widetilde{C}_2(T_0,E_{\MM},D_{\MM})$.
Then,
\begin{flalign*}
&\sum_{t=T_0}^{T-1}(t-T_0+1)(t-T_0+2)h_{\MM}(t,T_0)
%\nonumber\\
% &=8\eta^2(L+1)\sum_{t=0}^{T-T_0-1}(t+1)(t+2) p_c(t+T_0)\nonumber\\
% &+4\eta^2L^2\sum_{t=0}^{T-T_0-1}(t+1)(t+2)\gamma^2(t)p_c(t+T_0)\nonumber\\
% &+2\eta G(L+1)\hspace{-0.15cm}\sum_{t=0}^{T-T_0-1}\hspace{-0.15cm}(t+1)(t+2)\gamma(t)p_c(t+T_0)
\leq  C_2(T_0).
\end{flalign*}
We conclude the proof by letting 
\begin{flalign}\label{eq:def_C_MMT_0}
C_{\MM}(T_0)=2C_1(T_0)+\mu_{\epsilon} C_2(T_0).
\end{flalign}
%\my{Note that $C_{\MM}(T_0)$ captures the influence of the malicious agents through the terms (II)-(VI).}
% this had moved to the discussion immediately after the theorem
\end{proof}

Thus, we have shown that indeed we are able to recover convergence \emph{to the optimal value} of the original distributed optimization problem given in \eqref{eq:dist_opt_obj}, even in the presence of malicious agents. Further, we have established an upper bound on the expected value of $\|x_i(t)-x_{\LL}^{\star}\|^2$, for all $i\in\LL$, as a function of the time $t$ as given by~\eqref{eq:upper_bound_suboptimality_gap_alternative} in Theorem~\ref{theorem:exp_convergence_rate_alternative}.

\section{Numerical Results}\label{sec:numerical_results}
This section presents numerical results that validate the convergence results we derived for Algo.~\ref{alg:agent_i_dynamic}.

% As a benchmark, we compare the performance of our proposed Algo.~\ref{alg:agent_i_dynamic} to that of \cite{Sundaram_distributed_opt_malicious_2019} which adapts the  W-MSR consensus algorithm \cite{leBlancWMSR} to the case of distributed optimization. 
% This W-MSR based algorithm is applicable only for the one-dimensional case, i.e. $d=1$. To this end, we compare our results to \cite{W_MSR_Multi} which extends \cite{Sundaram_distributed_opt_malicious_2019} to multi-dimensional data values.

\begin{figure}[t!]
\vspace{-0.5cm}
    \centering
    \includegraphics[scale=0.75,trim={3.5cm 4cm 3.5cm 2.7cm},clip]{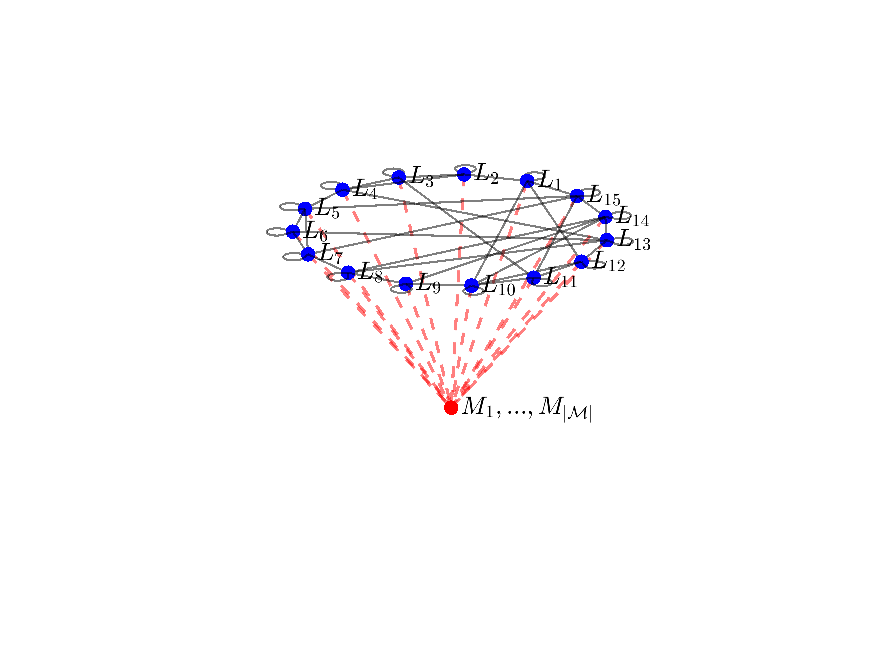}
    \vspace{-0.3cm}
    \caption{{\footnotesize Undirected graph $\mathbb{G}$. Two agents are neighbors if they are connected by an edge. Legitimate and malicious agents are depicted by blue and red nodes, respectively. Since every malicious agent is connected to all legitimate agents, all malicious agents are captured by a single node for simplicity of presentation. Edges between legitimate agents are depicted by black solid lines. Edges between legitimate and malicious agents are depicted by red dashed lines.}}
    \label{fig:connectivity_graph}
   % \vspace{-0.7cm}
\end{figure}

We consider a distributed network with $|\LL|=15$ legitimate agents and $|\MM|\in\{15,30\}$ malicious agents. To maximize the malicious agents'  impact, every malicious agent is connected to all the legitimate agents. The legitimate agent's connectivity is captured by Fig.~\ref{fig:connectivity_graph}. 

The trust values are generated as follows. Let $\EEop[\alpha_{ij}(t)] = 0.55$ if $j\in\mathcal{N}_i\cap \LL$, 
and $\EEop[\alpha_{ij}(t)] = 0.45$ if $j\in\mathcal{N}_i\cap \MM$. The random variable $\alpha_{ij}(t)$ is uniformly distributed on the interval $\left[\EEop[\alpha_{ij}(t)]-\frac{\ell}{2},\EEop[\alpha_{ij}(t)]+\frac{\ell}{2}\right]$, for every $i\in\LL$ and $j\in\LL\cup\MM$. We consider  the values $\ell$: $0.6,0.8$, in both scenarios $|E_{\LL}|=|E_{\MM}|=0.05$, however, the variance of the trust values when $\ell=0.8 $ are higher.  We remark that the legitimate agents are ignorant regarding the values $\EEop[\alpha_{ij}]$ and $\ell$. Due to the stochasticity of the trust values, we average the numerical results across $100$ system realizations. Furthermore, 
we use the following stepsize   $\gamma(t)=\frac{2}{(1-\epsilon)(t+2)}\cdot\mathds{1}_{\{t\geq 0\}},\:\: \epsilon=0.1$.

Denote for every client $i$,  $a_i\in\mathbb{R}^d, b_i\in\mathbb{R}$. 
Additionally, denote $\lambda\geq0$, and define for every agent $i\in\LL$ the following strongly convex loss with $\ell_2$  regularizer:
\[f_i(x)=\frac{1}{2}(a_i^Tx-b_i)^2+\frac{\lambda}{2}\|x\|^2.\]
We constrain the legitimate agents' values to lie in the $d$-dimensional ball that is centered at the origin,  i.e., $\XX=\{x\in\mathbb{R}^d:\|x\|\leq \eta\}$, where $\eta=30$. 
It follows that the global optimization problem the legitimate agents aim to solve distributively is
\begin{flalign}\label{eq:min_general_num_results}
    \min_{\|x\|\leq\eta} \left\{\frac{1}{|\LL|}\sum_{i\in\LL}\frac{1}{2}(a_i^Tx-b_i)^2+\frac{\lambda}{2}\|x\|^2\right\}.
\end{flalign}

Furthermore, denote by $x^{\star \text{UC}}_{\LL}$ the optimal point of the unconstrained counterpart of \eqref{eq:min_general_num_results}, that is
$x^{\star\text{UC}}_{\LL}=\left(\lambda I+\frac{1}{|\LL|}\sum_{i\in\LL}a_ia_i^T\right)^{-1}\left(\frac{1}{|\LL|}\sum_{i\in\LL}a_i b_i\right)$.
Then, the optimal point of \eqref{eq:min_general_num_results} is
\[x^{\star}_{\LL}=\left((\nu+\lambda) I+\frac{1}{|\LL|}\sum_{i\in\LL}a_ia_i^T\right)^{-1}\left(\frac{1}{|\LL|}\sum_{i\in\LL}a_i b_i\right),\]
where $\nu$ is a dual variable that is set to zero if $\|x^{\star\text{UC}}_{\LL}\|<\eta$ 
 Otherwise, it is set such that $\|x^{\star}_{\LL}\|=\eta$.

In this setup for every $i\in\LL$ and $x\in\mathbb{R}^d$ 
\[\nabla f_i(x)= a_i(a_i^Tx-b_i)+\lambda x \text{ and } \Pi_{\XX}\left(x\right)= \min\left\{1,\frac{\eta}{\|x\|}\right\}\cdot x.\]
To evaluate the performance of Algo.~\ref{alg:agent_i_dynamic} we denote the average squared error at time $t$ by $\overline{e}(t)\triangleq\frac{1}{|\LL|}\sum_{i\in\LL}\|x_i(t)-x_{\LL}^{\star}\|^2$. 

The W-MSR algorithm \cite{Sundaram_distributed_opt_malicious_2019} for robust distributed optimization, which we use for comparison in the supplementary material \cite{our_arXiv_with_supplementary} for the one-dimensional case, is only valid for one-dimensional data values. Thus, as a benchmark we compare our results to the multi-dimensional extension of the W-MSR algorithm proposed in \cite{W_MSR_Multi}. 
Following the notations in \cite{Sundaram_distributed_opt_malicious_2019}, we denote by $F$ the maximal number of highest values and lowest values that each legitimate agent discards, overall a legitimate agent may ignore no more than $2F$ values for $d=1$. We note that in this case the number of tolerated malicious inputs for the W-MSR algorithm is reduced by a factor of $d$, i.e., the dimension of the data values. Additionally, we use the nominal case with no malicious agents \eqref{eq:no_malicious_update_opt_gradient_selfish_analysis_simple} as another benchmark to evaluate the performance of Algo. \ref{alg:agent_i_dynamic}.

In our numerical results, we use the following auxiliary vector 
$(\tilde{b}_i)_{i=1}^{15}=(115.7, 163.3, -81.7, 127.2,-63.7,58.4,-3.1,$ $   62.9,
 54.5, 144.9, -121.1, 9.3, -2.6, -124.5, 131)$.
Note that setting $d=1$, $a_i=1$, and $\lambda=0$ results in a one-dimensional constrained consensus problem. We consider this special case in Section~III in the supplementary material \cite{our_arXiv_with_supplementary}. 

Next, we examine a multi-dimensional setup where both the optimal solution and the updates are affected by the constraint set $\XX$.
We set $d=5$ and $\lambda=0.1$. Additionally, we set $b_i=\tilde{b}_i/2$ for all $i\in\LL$, and
 \begin{flalign}
  \hspace{-0.2cm}  \left[(a_i)_{i=1}^{15}\right]^T ={\footnotesize\begin{pmatrix}
-0.87 & -1.05 & -2.81 & -0.4 & -1.76\\
-0.88 & -0.34 & 0.34 & -2.46 & 0.44\\
-0.25 & 0.47 & -0.09 & -0.99 & -2.33\\
-0.27 & -0.61 & -2.5 & -0.79 & 0.46\\
-0.23 & 1.83 & 0.89 & -0.83 & -0.67\\
-1.6 & 0.27 & -0.81 & -2.77 & -0.21\\
-1.42 & -1.11 & -1.63 & -0.66 & -1.54\\
-1.19 & -0.3 & -1.97 & -1.42 & -1.21\\
-1.43 & -1.64 & 0.17 & -2.11 & -2.11\\
-0.73 & 0.46 & -0.42 & -1.75 & 0.22\\
-0.97 & -0.12 & -2.35 & -2.51 & -1.63\\
-1.18 & -1.42 & -0.13 & -1.66 & 0.36\\
-0.63 & -2.19 & -1.15 & -1.65 & -2.02\\
 0.59 & -2.08 & 0.26 & -0.74 & -2.66\\
-3.05 & -0.7 & 0.2 & -1.94 & -1.4
    \end{pmatrix}}.
\end{flalign}

The optimal point of the nominal dynamic~\eqref{eq:no_malicious_update_opt_gradient_selfish_analysis_simple}, when no malicious agents are present, rounded to the second digit after the decimal point is
$x_{\LL}^{\star} \approx (-20.95, -6.15, -5.92, -3.54, 19.39)^T$. Additionally, the inputs of the malicious agents at all times are $-x_{\LL}^{\star}$ which is the farthest point from $x_{\LL}^{\star}$ in $\XX$.

\begin{figure}[t!]
    \centering
    \vspace{-0.5cm}
    \includegraphics[scale=0.55,trim={0.05cm 0cm 0cm 0.03cm},clip]{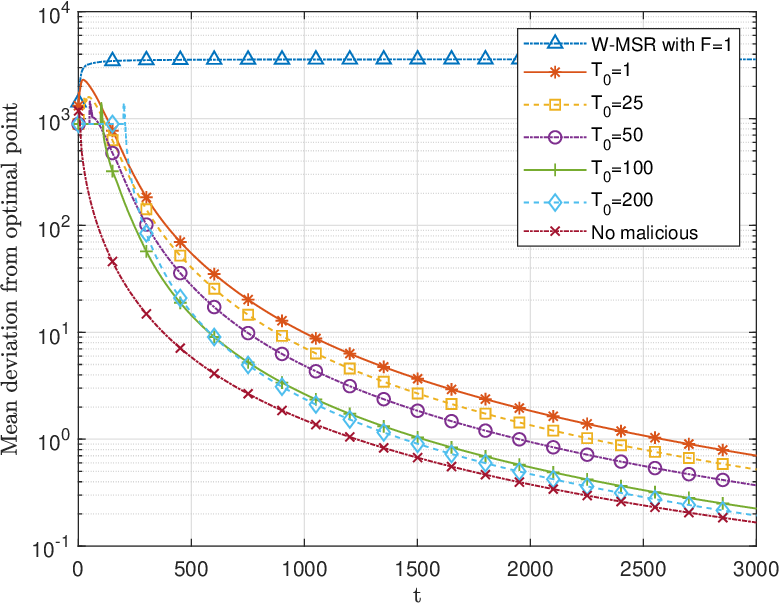}
        \vspace{-0.3cm}
    \caption{{\small Average  $\overline{e}(t)$ as a function of $t$ for $|\MM|=15,\:\ell=0.8$.}}
    \label{fig:fig_15_malicious5D}
\vspace{0.3cm}
    \centering
    \includegraphics[scale=0.55,trim={0.05cm 0cm 0cm 0.03cm},clip]{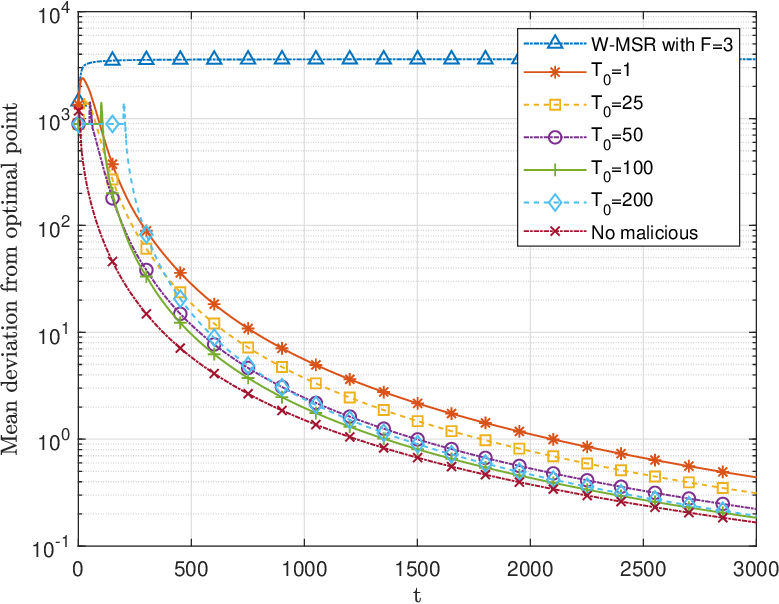}
    \vspace{-0.3cm}
    \caption{{\small Average  $\overline{e}(t)$ as a function of $t$ for $|\MM|=30,\:\ell=0.6$.}}
    \label{fig:fig_30_malicious5D}
   % \vspace{-0.7cm}
\end{figure}

Figs.~\ref{fig:fig_15_malicious5D} and \ref{fig:fig_30_malicious5D} capture the average value of the squared distance of each legitimate agent from the optimal point $x_{\LL}^{\star}$, i.e., the average value of $\overline{e}(t)$, for each time $t$.
The plots presented in Figs.~\ref{fig:fig_15_malicious5D} and~\ref{fig:fig_30_malicious5D} for our higher dimensional setup are consistent with the one-dimensional setup which is captured in Figs.~1-4 in the supplementary material \cite{our_arXiv_with_supplementary}.  
We can see that Algo.~\ref{alg:agent_i_dynamic} performs well and mitigates the harmful effect of malicious inputs even in higher dimensions. This is in contrast to the multi-dimensional W-MSR algorithm \cite{W_MSR_Multi} which is more vulnerable to malicious attacks as the dimension of the data values, i.e., $d$, increases.

In particular, Algo.~\ref{alg:agent_i_dynamic} provides resilience to malicious activity and can tolerate even $30=2|\LL|$ malicious agents, as evident  in Figs.~\ref{fig:fig_15_malicious5D} and~\ref{fig:fig_30_malicious5D}. Furthermore, we can see from Figs.~\ref{fig:fig_15_malicious5D} and \ref{fig:fig_30_malicious5D} that Algo.~\ref{alg:agent_i_dynamic} is robust to undesirably small values of $|E_{\LL}|$ and $|E_{\MM}|$ where the trust observations are less informative. 
Figs.~\ref{fig:fig_15_malicious5D} and~\ref{fig:fig_30_malicious5D} demonstrate how our algorithm eventually recovers the global optimum as predicted by theory regardless of the value of the observation window $T_0$. Thus, the value of $T_0$ mostly dictates \emph{the rate} of recovery of the global optimum.
Particularly, Figs.~\ref{fig:fig_15_malicious5D} and~\ref{fig:fig_30_malicious5D} show that the variance of the trust values has more impact on $T_0$ values that are smaller than $100$. This occurs since the higher variance of the trust values increases the misclassification errors. 
Since the probability of these errors decreases with $T_0$, they are less impactful when $T_0$ is  $100$ or higher. Finally, we refer the reader to Section~III in the supplementary material \cite{our_arXiv_with_supplementary}  for additional numerical results. 

\section{Conclusions}\label{sec:conclusions}
This work studies the problem of resilient distributed optimization in the presence of malicious activity. We consider the case where additional information in the form of stochastic inter-agent trust values is available. Under this model, we propose a mechanism for exploiting these trust values where legitimate agents learn to distinguish between their legitimate and malicious neighbors. We incorporate this mechanism to arrive at resilient distributed optimization where strong performance guarantees can be recovered. Specifically, we prove that our algorithm converges to the optimal solution of the nominal distributed optimization system with no malicious agents, both in expectation and almost surely.
Additionally, we present two mechanisms to derive upper bounds on the expected distance of the agents' iterates from the optimal solution. The first is based on the correct classification of all malicious and legitimate agents. The second approach utilizes the dynamic of the attacked system to carefully tighten the upper bound on the expected convergence rate. Finally, we present numerical results that demonstrate the performance of our proposed distributed optimization framework.

The use of stochastic trust values allows us to recover convergence to the \emph{global optimum} in distributed optimization problems even when more than half of the network is malicious. This represents a very challenging case where not many strong performance results are currently available, particularly in the case of distributed optimization problems. Thus the results of this paper strengthen our characterization of achievable performance and provide novel tools for resilient trust-centered optimization in multi-agent systems.

% \vspace{-0.1cm}
\appendices
% \vspace{-0.2cm}
\input{append_classification_prob}
% \vspace{-0.2cm}
\input{append_convergence_in_mean_proof_definition}

% \vspace{-0.2cm}
\input{append_distance_mean}
% \vspace{-0.2cm}


%%%%%%%%%%%%%%%%%%%%%%%%%%%%%%%%%%%%%%%%%%%%%%%%%%%%%%%%%%%%%%%%%%%%%%%%%%%%%%%%
%\addtolength{\textheight}{-1cm} 
%\vspace{-0.5cm}

\section*{References}
%\vspace{-0.7cm}

\bibliographystyle{IEEEtran}
%\bibliographystyle{alpha}

\end{document}

\title{Supplementary Material:\\
Resilient Distributed Optimization for \\
Multi-Agent Cyberphysical Systems}
\author{Michal Yemini, Angelia Nedi\'c, Andrea J. Goldsmith, Stephanie Gil%
\thanks{}
}

\maketitle
\input{commands}
For the sake of completeness of presentation, this supplementary material includes complimentary proofs and discussions that the main manuscript could not elaborate on. Specifically, Section \ref{sec_sup:proof_nominal} presents the proof of Theorem \ref{theorem:convergence_no_malicious} which analyzes the dynamic of the nominal case with no malicious agents. Section \ref{sec_sup:proof_theorem_gap} presents an alternative proof by definition of Theorem \ref{theorem:expected_suboptimality_gap}. Finally, Section \ref{sec_sup:additional_numerical} includes additional numerical results and evaluations.

\section{Proof of Theorem \ref{theorem:convergence_no_malicious}}\label{sec_sup:proof_nominal}
 Here, we aim to explore the dynamic \eqref{eq:legitimate_update_opt_gradient_selfish} when $T_0=0$, $\MM=\emptyset$ and the legitimate agents are aware that all the agents are legitimate. In this scenario the dynamic \eqref{eq:legitimate_update_opt_gradient_selfish} is equivalent to 
        \begin{flalign}\label{eq:no_malicious_update_opt_gradient_selfish_analysis}
    r_{i}(t)&=\overline{w}_{ii} z_i(t)+ \sum_{j\in\NN_i\cap\LL}\overline{w}_{ij}z_j(t)\nonumber\\
    y_{i}(t)&=r_{i}(t)-\gamma(t)\nabla f_i(r_{i}(t)),\nonumber\\
    \phi_{i}(t)&=\Pi_{\XX}\left(y_i(t)\right)-r_{i}(t),\nonumber\\
    z_i(t+1) &=  r_{i}(t)+\phi_{i}(t)=\Pi_{\XX}\left(y_i(t)\right).
    \end{flalign}
    Denote $Z(t)\triangleq (z_1(t),\ldots,z_{|\LL|}(t))$ and $\Phi(t)\triangleq (\phi_1(t),\ldots,\phi_{|\LL|}(t))$, then 
    \[Z(t+1)=\overline{W}_{\LL}Z(t)+\Phi(t),\]
    where by Lemma \ref{lemma:upper_bound_phi}, $\|\phi_i(t)\|\leq \gamma(t)G$.
    Thus we can utilize Proposition \ref{prop:distance_to_mean_time_t} to upper bound the distance of an agent value from that of the average.      Denote \[\overline{z}(t)\triangleq\frac{1}{|\LL|}\sum_{i\in\LL}z_i(t),\] and  
    \begin{flalign}
    g(t) \triangleq \min\left\{2\eta,\rho_{\LL}^{t}2\eta+\frac{\rho_{\LL}^{t/2}G\gamma(0)}{1-\rho_{\LL}}+\frac{G\gamma(t/2)}{{1-\rho_{\LL}}}\right\}.
    \end{flalign}
    Observe that by the double stochasticity of $\overline{W}_{\LL}$ we have that $\overline{z}(t)=\frac{1}{|\LL|}\sum_{i\in\LL}r_i(t)$.
    \begin{corollary}\label{corollary:dstance_from_avg}
    For every  $t\in\mathbb{|\LL|}$
    \begin{flalign}
    \frac{1}{|\LL|}\sum_{i\in\LL}\|z_i(t)-\overline{z}(t)\|&\leq g(t),\text{ and }\nonumber\\
    \frac{1}{|\LL|}\sum_{i\in\LL}\|z_i(t)-\overline{z}(t)\|^2&\leq g^2(t).
    \end{flalign}
    \end{corollary}
    \begin{proof}
    This is a direct result of \eqref{eq:max_val_norm} and of Proposition \ref{prop:distance_to_mean_time_t} (in the main manuscript), for deterministic $\Delta(t)$ where 
    \[[\Delta(t)]_i=\phi_i(t),\]
    and
    \[\delta(t)=\gamma(t)G.\]
    \end{proof}
    We are now ready to prove Theorem \ref{theorem:convergence_no_malicious}.
    \begin{proof}[Proof of Theorem \ref{theorem:convergence_no_malicious}]
    Under the dynamic \eqref{eq:no_malicious_update_opt_gradient_selfish_analysis}, for every $i\in\LL$ and $t\in\mathbb{|\LL|}$
    \begin{flalign*}
    &\frac{1}{|\LL|}\sum_{i\in\LL}\|z_i(t+1)-x_{\LL}^{\star}\|^2\nonumber\\
    &=\frac{1}{|\LL|}\sum_{i\in\LL}\|\Pi_{\XX}(y_i(t))-x_{\LL}^{\star}\|^2\nonumber\\
     &=\frac{1}{|\LL|}\sum_{i\in\LL}\|\Pi_{\XX}(y_i(t))-\Pi_{\XX}(x_{\LL}^{\star})\|^2\nonumber\\
    &\leq\frac{1}{|\LL|}\sum_{i\in\LL}\|y_i(t)-x_{\LL}^{\star}\|^2\nonumber\\
    &=\frac{1}{|\LL|}\sum_{i\in\LL}\|r_{i}(t)-\gamma(t)\nabla f_i(r_{i}(t))-x_{\LL}^{\star}\|^2\nonumber\\
    &=\frac{1}{|\LL|}\sum_{i\in\LL}\|r_{i}(t)-x_{\LL}^{\star}\|^2+\frac{1}{|\LL|}\gamma^2(t)\sum_{i\in\LL}\|\nabla f_i(r_{i}(t))\|^2-\frac{2}{|\LL|}\gamma(t)\sum_{i\in\LL}[\nabla f_i(r_{i}(t))]^T(r_{i}(t)-x_{\LL}^{\star}).
    \end{flalign*}
    The convexity of $\|\cdot\|^2$  and the double stochasticity of $\overline{W}_{\LL}$ yields that
    \begin{flalign*}
    \frac{1}{|\LL|}\sum_{i\in\LL}\|r_{i}(t)-x_{\LL}^{\star}\|^2&\leq \frac{1}{|\LL|}\sum_{i\in\LL}\|z_i(t)-x_{\LL}^{\star}\|^2,
    \end{flalign*}
    and 
    \begin{flalign*}
     \|\overline{z}(t)-x_{\LL}^{\star}\|^2&=\left\|\frac{1}{|\LL|}\sum_{i\in\LL}r_{i}(t)-x_{\LL}^{\star}\right\|^2\leq  \frac{1}{|\LL|}\sum_{i\in\LL}\|r_{i}(t)-x_{\LL}^{\star}\|^2.
    \end{flalign*}
    Additionally, we can conclude from Corollary \ref{corollary:bounded_gradients} that
    \begin{flalign*}
    \frac{1}{|\LL|}\gamma^2(t)\sum_{i\in\LL}\|\nabla f_i(r_{i}(t))\|^2\leq
    \gamma^2(t)G^2.
    \end{flalign*}
    Thus, by the $\mu$-strong convexity of $f_i,\:\forall\: i\in\LL$
    \begin{flalign}\label{eq:upper_inner_dynamic_no_malicious}
    &-2\frac{\gamma(t)}{|\LL|}\sum_{i\in\LL}[\nabla f_i(r_{i}(t))]^T(r_{i}(t)-x_{\LL}^{\star})
    \nonumber\\
    &=\frac{2\gamma(t)}{|\LL|}\sum_{i\in\LL}[\nabla f_i(r_{i}(t))]^T(x_{\LL}^{\star}-r_{i}(t))\nonumber\\
    &\leq \frac{2\gamma(t)}{|\LL|}\sum_{i\in\LL}\left[f_i(x_{\LL}^{\star})-f_i(r_{i}(t))-\frac{\mu}{2}\|r_{i}(t)-x_{\LL}^{\star}\|^2\right]\nonumber\\
     &\leq \frac{2\gamma(t)}{|\LL|}\sum_{i\in\LL}\left[f_i(x_{\LL}^{\star})-f_i(r_{i}(t))\right]-\frac{2\gamma(t)\mu}{2}\|\overline{z}(t)-x_{\LL}^{\star}\|^2\nonumber\\
    &= \underbrace{\frac{2\gamma(t)}{|\LL|}\sum_{i\in\LL}\left[f_i(x_{\LL}^{\star})-f_i(\overline{z}(t))\right]}_{\leq 0}+\frac{2\gamma(t)}{|\LL|}\sum_{i\in\LL}[f_i(\overline{z}(t))-f_i(r_{i}(t))]-\gamma(t)\mu\|\overline{z}(t)-x_{\LL}^{\star}\|^2.
    \end{flalign}

    Now, we upper bound the term $-\gamma(t)\mu\|\overline{z}(t)-x_{\LL}^{\star}\|^2$ as follows:
    \begin{flalign}
    -\gamma(t)\mu\|\overline{z}(t)-x_{\LL}^{\star}\|^2&= -\frac{\gamma(t)\mu}{|\LL|}\sum_{i\in\LL}\|\overline{z}(t)-z_i(t)+z_i(t)-x_{\LL}^{\star}\|^2\nonumber\\
    &= -\frac{\gamma(t)\mu}{|\LL|}\sum_{i\in\LL}\|\overline{z}(t)-z_i(t)\|^2-\frac{\gamma(t)\mu}{|\LL|}\sum_{i\in\LL}\|z_i(t)-x_{\LL}^{\star}\|^2- \frac{2\gamma(t)\mu}{|\LL|}\sum_{i\in\LL}(\overline{z}(t)-z_i(t))^T(z_i(t)-x_{\LL}^{\star})\nonumber\\
    &\stackrel{(*)}{\leq}-\frac{\gamma(t)\mu}{|\LL|}\sum_{i\in\LL}\|\overline{z}(t)-z_i(t)\|^2-\frac{\gamma(t)\mu}{|\LL|}\sum_{i\in\LL}\|z_i(t)-x_{\LL}^{\star}\|^2+ \frac{2\gamma(t)\mu}{|\LL|}\sum_{i\in\LL}\|\overline{z}(t)-z_i(t)\|\cdot\|z_i(t)-x_{\LL}^{\star}\|\nonumber\\
    \end{flalign}
    where (*) follows from the Cauchy-Schwarz.

Let $\epsilon\in(0,1)$, then by Young inequality we have that 
\[\|\overline{z}(t)-z_i(t)\|\cdot\|z_i(t)-x_{\LL}^{\star}\|\leq \frac{1}{2\epsilon}\cdot\|\overline{z}(t)-z_i(t)\|^2+\frac{\epsilon}{2}\cdot\|z_i(t)-x_{\LL}^{\star}\|^2.\]
Thus, we have that
\begin{flalign}
-\gamma(t)\mu\|\overline{z}(t)-x_{\LL}^{\star}\|^2&\leq 
-\frac{\gamma(t)\mu(1-\epsilon)}{|\LL|}\sum_{i\in\LL}\|z_i(t)-x_{\LL}^{\star}\|^2+\frac{\gamma(t)\mu(1/\epsilon-1)}{|\LL|}\sum_{i\in\LL}\|\overline{z}(t)-z_i(t)\|^2.
\end{flalign}

    Next, we upper bound the term $\frac{2\gamma(t)}{|\LL|}\sum_{i\in\LL}[f_i(\overline{z}(t))-f_i(r_{i}(t))]$ in  \eqref{eq:upper_inner_dynamic_no_malicious} by utilizing the $L$ Lipschitz continuity of $\nabla f_i$:
    \begin{flalign*}
    &\frac{2\gamma(t)}{|\LL|}\sum_{i\in\LL}[f_i(\overline{z}(t))-f_i(r_{i}(t))]\nonumber\\
    &\leq  \frac{2\gamma(t)}{|\LL|}\sum_{i\in\LL}\bigg[(\nabla f_i(r_{i}(t)))^{T}(\overline{z}(t)-r_{i}(t))+\frac{L}{2}\|r_i(t)-\overline{z}(t)\|^2\bigg]\nonumber\\
    &\stackrel{(*)}{\leq}  \frac{2\gamma(t)}{|\LL|}\sum_{i\in\LL}\bigg[(\nabla f_i(r_{i}(t)))^{T}(\overline{z}(t)-r_{i}(t))+\frac{L}{2}\|z_i(t)-\overline{z}(t)\|^2\bigg]\nonumber\\
    &\leq \frac{2\gamma(t)}{|\LL|}\sum_{i\in\LL}\left[G\|r_{i}(t)-\overline{z}(t)\|+\frac{L}{2}\|z_i(t)-\overline{z}(t)\|^2\right]\nonumber\\
    &\leq \frac{2\gamma(t)}{|\LL|}\sum_{i\in\LL}\left[G\|z_i(t)-\overline{z}(t)\|+\frac{L}{2}\|z_i(t)-\overline{z}(t)\|^2\right].
    \end{flalign*}
where $(*)$ follows from the double stochasticity of $\overline{W}$.
    
    It follows that, 
    \begin{flalign}\label{eq:main_recursion_no_malicious_analysis}
    &\frac{1}{|\LL|}\sum_{i\in\LL}\|z_i(t+1)-x_{\LL}^{\star}\|^2\nonumber\\
    &\leq \frac{1-\gamma(t)\mu(1-\epsilon)}{|\LL|}\sum_{i\in\LL}\|z_i(t)-x_{\LL}^{\star}\|^2+\gamma^2(t)G^2+\frac{2\gamma(t)}{|\LL|}\sum_{i\in\LL}\left[G\|z_i(t)-\overline{z}(t)\|+\frac{\mu(1/\epsilon-1)+L}{2}\|z_i(t)-\overline{z}(t)\|^2\right].
    \end{flalign}

Denote, 
\[h_{\epsilon}(t)=G^2\gamma(t)+2Gg(t)+(\mu(1/\epsilon-1)+L)g^2(t).\]
   We  use  Corollary (Supplementary) \ref{corollary:dstance_from_avg} to deduce that
    \begin{flalign}
    &\frac{1}{|\LL|}\sum_{i\in\LL}\|z_i(t+1)-x_{\LL}^{\star}\|^2
    \leq
    \left(1-\mu\gamma(t)(1-\epsilon)\right)\frac{1}{|\LL|}\sum_{i\in\LL}\|z_i(t)-x_{\LL}^{\star}\|^2
    +\gamma(t)h_{\epsilon}(t).
    \end{flalign}
Consequently,
\begin{flalign}
\lim_{t\rightarrow\infty}\|z_i(t)-x_{\LL}^{\star}\|= 0,\:\forall \: i\in\LL,
\end{flalign}
whenever $\sum_{t=0}^{\infty}\gamma(t)=\infty$ and $\sum_{t=0}^{\infty}\gamma^2(t)<\infty$.

Motivated by \cite{O_1_t_centralize_strong_convex} we let $\gamma(t)=\frac{2}{\mu_{\epsilon}(t+2)}$ where $\mu_{\epsilon}\triangleq\mu(1-\epsilon)$. 
It follows that
    \begin{flalign}
    &\frac{1}{|\LL|}\sum_{i\in\LL}\|z_i(t+1)-x_{\LL}^{\star}\|^2
 \leq
    \frac{t}{t+2}\cdot\frac{1}{|\LL|}\sum_{i\in\LL}\|z_i(t)-x_{\LL}^{\star}\|^2
    +\frac{2h_{\epsilon}(t)}{\mu_{\epsilon}(t+2)}.
    \end{flalign}
Multiplying both sides by $(t+1)(t+2)$ yields the following upper bound
    \begin{flalign}
    &(t+1)(t+2)\frac{1}{|\LL|}\sum_{i\in\LL}\|z_i(t+1)-x_{\LL}^{\star}\|^2
    \leq
    t(t+1)\cdot\frac{1}{|\LL|}\sum_{i\in\LL}\|z_i(t)-x_{\LL}^{\star}\|^2
    +\frac{2(t+1)h_{\epsilon}(t)}{\mu_{\epsilon}}.
    \end{flalign}
Summing both sides over the set $t={0,1,\ldots,T-1}$ yields the upper bound:
    \begin{flalign}
    &\frac{1}{|\LL|}\sum_{i\in\LL}\|z_i(T)-x_{\LL}^{\star}\|^2
    \leq \frac{2\sum_{t=0}^{T-1}(t+1)h_{\epsilon}(t)}{\mu_{\epsilon} T(T+1)}.
    \end{flalign}
    To conclude the proof we upper bound the term $\sum_{t=0}^{T-1}(t+1)h(t)$. 
     To this end, we utilize the identity \[\sum_{t=0}^{\infty}(t+1)x^t=(1-x)^{-2},\] 
     for all $|x|\in(0,1)$ as follows
    \begin{flalign}
    &\sum_{t=0}^{T-1}(t+1)h_{\epsilon}(t)= \sum_{t=0}^{T-1}(t+1)\left(G^2\gamma(t)+2Gg(t)+(\mu_{\epsilon}/\epsilon+L)g^2(t)\right).
    \end{flalign}
    Now,
    \begin{flalign}
     \sum_{t=0}^{T-1}(t+1)G^2\gamma(t)=\sum_{t=0}^{T-1}(t+1)G^2\frac{2}{\mu_{\epsilon}(t+2)}\leq \frac{2G^2T}{\mu_{\epsilon}}.
    \end{flalign}
Additionally, 
\begin{flalign}
 \sum_{t=0}^{T-1}(t+1)2Gg(t)
 &\leq 2G\sum_{t=0}^{T-1}(t+1)\left[\rho_{\LL}^{t}2\eta+\frac{\rho_{\LL}^{t/2}G\gamma(0)}{1-\rho_{\LL}}+\frac{G\gamma(t/2)}{{1-\rho_{\LL}}}\right]\nonumber\\
 &\leq 4\eta G\sum_{t=0}^{T-1}(t+1)\rho_{\LL}^{t}+\frac{2G^2\gamma(0)}{1-\rho_{\LL}}\sum_{t=0}^{T-1}(t+1)\rho_{\LL}^{t/2}+\frac{8G^2T}{\mu_{\epsilon}(1-\rho_{\LL})}\nonumber\\
 &\stackrel{(*)}{\leq} \frac{4\eta G}{(1-\rho_{\LL})^2}+\frac{8G^2\gamma(0)}{1-\rho_{\LL}}\sum_{t=0}^{T-1}(t+1)\rho_{\LL}^{t}+\frac{8G^2T}{\mu_{\epsilon}(1-\rho_{\LL})}\nonumber\\
  &\stackrel{(**)}{=} \frac{4\eta G}{(1-\rho_{\LL})^2}+\frac{8G^2}{\mu_{\epsilon}(1-\rho_{\LL})^3}+\frac{8G^2T}{\mu_{\epsilon}(1-\rho_{\LL})},
\end{flalign}
where inequality $(*)$ follows since 
\begin{flalign}
 \sum_{t=0}^{T-1}(t+1)\rho_{\LL}^{t/2}&\leq
 \sum_{t=0}^{T-1}(2t+1)\rho_{\LL}^{t}+\sum_{t=0}^{T-1}(2t+2)\rho_{\LL}^{t}
 \leq 4\sum_{t=0}^{T-1}(t+1)\rho_{\LL}^{t},
\end{flalign}
and equality $(**)$ follows since $\gamma(0)=\frac{2}{\mu_{\epsilon}(0+2)}=\frac{1}{\mu_{\epsilon}}$
%Since $|\rho_{\LL}|<1$  then $(1-\rho_{\LL}^x)^{-1}\leq (1-\rho_{\LL})^{-1}$ for all $x\geq1$. It follows that

Additionally,
\begin{flalign}
 &(\mu_{\epsilon}/\epsilon+L)\sum_{t=0}^{T-1}(t+1)g^2(t)\nonumber\\
 &\leq (\mu(1/\epsilon-1)+L)\sum_{t=0}^{T-1}(t+1)\left[\rho_{\LL}^{t}2\eta+\frac{\rho_{\LL}^{t/2}G\gamma(0)}{1-\rho_{\LL}}+\frac{G\gamma(t/2)}{{1-\rho_{\LL}}}\right]^2\nonumber\\
 &\leq (\mu_{\epsilon}/\epsilon+L)\bigg[\frac{4\eta^2}{(1-\rho_{\LL}^2)^2}+\frac{4G^2}{\mu_{\epsilon}^2(1-\rho_{\LL})^2(1-\rho_{\LL}^2)^2}+\frac{16G^2\ln\left(\frac{T+2}{2}\right)}{\mu_{\epsilon}^2(1-\rho_{\LL})^2}\nonumber\\
 &\hspace{3cm}+\frac{16G\eta}{\mu_{\epsilon}(1-\rho_{\LL})(1-\rho_{\LL}^3)^2}+\frac{16G\eta}{\mu_{\epsilon}(1-\rho_{\LL})^2}+\frac{16G^2}{\mu_{\epsilon}^2(1-\rho_{\LL})^3}\bigg].
 %\nonumber\\&= \frac{4(\mu\eta+2G)^2}{\mu^2(1-\rho_{\LL})^2}+\frac{2G^2}{\mu^2(1-\rho_{\LL})^4}+\frac{8G\eta}{\mu(1-\rho_{\LL})^3}+ \frac{16G^2}{\mu^2(1-\rho_{\LL})^2}\ln\left(\frac{T+2}{2}\right).
\end{flalign}
Finally,  by utilizing the inequality $(1-x^k)^{-1}\leq (1-x)^{-1}. \:\forall x\in[0,1)$, and the notation \eqref{eq:h_overline_T_def} (in the main manuscript)
we can conclude that
 \[\sum_{t=0}^{T-1}(t+1)h(t)\leq 2\overline{h}_{\epsilon}(T).\]
\end{proof}
We note that it may be possible to extend the explicit upper bound presented in Theorem \ref{theorem:convergence_no_malicious} to a more general choice of $\gamma_t$ by extending the recently proposed \cite[Lemma 6]{li2022multi}. This can lead to tighter upper bounds where an optimized stepsize can be chosen to reduce the current upper bound.

\section{Proof by Definition of Theorem \ref{theorem:expected_suboptimality_gap}}\label{sec_sup:proof_theorem_gap}
For the sake of completeness of discussion, we provide an alternative proof for Theorem \ref{theorem:expected_suboptimality_gap}, where we use the law of total expectation with respect to the events $\{T_f=k\}_{k=0}^{t-1}$ and $\{T_f>t-1\}$.
\begin{proof}[An Alternative Proof for Theorem \ref{theorem:expected_suboptimality_gap}]
First, note that $\frac{\overline{h}_{\epsilon}(t)}{t(t+1)}$ is nonincreasing function of $t$ for $t\geq1$ and $
\mu_{\epsilon}>0$. Denote $M_k\triangleq\max\{k,T_0\}$.
For every $t-1\geq T_0$ and $m\in [T_0:t-1]$ we have that
\begin{flalign*}
&\frac{1}{|\LL|}\sum_{i\in\LL}\EEop\left[\|x_{i}(t)-x_{\LL}^{\star}\|^2\right]\nonumber\\
&\stackrel{(a)}{=} \frac{1}{|\LL|}\sum_{k=0}^{t-1}\Pr(T_f= k)\sum_{i\in\LL}\EEop\left[\|x_{i}(t)-x_{\LL}^{\star}\|^2\big|T_f=k\right]+\Pr(T_f>t-1)\frac{1}{|\LL|}\sum_{i\in\LL}\EEop\left[\|x_{i}(t)-x_{\LL}^{\star}\|^2\big|T_f>t-1\right]\nonumber\\
&\stackrel{(b)}{\leq} \frac{1}{|\LL|}\sum_{k=0}^{t-1}\sum_{i\in\LL}\Pr(T_f= k)\EEop\left[\|z_i(t-M_k)-x_{\LL}^{\star}\|^2\:\big|T_f=k\right]+4\eta^2\Pr(T_f>t-1)\nonumber\\
&\stackrel{(c)}{\leq} \sum_{k=0}^{t-1}\Pr(T_f= k)\min\left\{4\eta^2,\frac{4\overline{h}_{\epsilon}(t-M_k)}{\mu_{\epsilon}(t-M_k)(t-M_k+1)}\right\}+4\eta^2\Pr(T_f>t-1)\nonumber\\
& \leq \sum_{k=0}^{t-1} \Pr(T_f= k)\min\left\{4\eta^2,\frac{4\overline{h}_{\epsilon}(t-M_k)}{\mu_{\epsilon}(t-M_k)(t-M_k+1)}\right\}+4\eta^2 \Pr(T_f>t-1)\nonumber\\
& \leq\sum_{k=0}^{m} \Pr(T_f= k)\min\left\{4\eta^2,\frac{4\overline{h}_{\epsilon}(t-M_k)}{\mu_{\epsilon}(t-M_k)(t-M_k+1)}\right\}\nonumber\\
&+\sum_{k=m+1}^{t-1}\Pr(T_f=k)\min\left\{4\eta^2,\frac{4\overline{h}_{\epsilon}(t-M_k)}{\mu_{\epsilon}(t-M_k)(t-M_k+1)}\right\}+4\eta^2 \Pr(T_f>t-1)\nonumber\\
& \leq \min\left\{4\eta^2,\frac{4\overline{h}_{\epsilon}\left(t-m\right)}{\mu_{\epsilon}(t-m)(t-m+1)}\right\}\underbrace{\sum_{k=0}^{m}\Pr(T_f= k)}_{\leq 1}+4\eta^2\sum_{k=m+1}^{t-1}\Pr(T_f=k)+4\eta^2 \Pr(T_f>t-1)\nonumber\\
& \leq \min\left\{4\eta^2,\frac{4\overline{h}_{\epsilon}\left(t-m\right)}{\mu_{\epsilon}(t-m)(t-m+1)}\right\}+4\eta^2\Pr(T_f>m)\nonumber\\
& = \min\left\{4\eta^2,\frac{4\overline{h}_{\epsilon}\left(t-m\right)}{\mu_{\epsilon}(t-m)(t-m+1)}\right\}+4\eta^2 p_e(m),
\end{flalign*}
where $(a)$ follows from the law of total expectation, $(b)$ follows from the definition of $T_f$ and Assumption \ref{assumption:domain_sets}, $(c)$ follows from Theorem \ref{theorem:convergence_no_malicious}, and the remaining steps follow 
since $\frac{\overline{h}_{\epsilon}(t)}{\mu_{\epsilon}t(t+1)}$ is nonincreasing function of $t$ for $t\geq1$ and from Lemma \ref{lemma:classification_prob}.
\end{proof}

\section{Additional Numerical Results}\label{sec_sup:additional_numerical}

In this section we present additional numerical results for the one-dimensional case. We first present numerical results of the mean squared error, and then include a numerical comparison of the upper bound for the average mean squared error and discuss their merits.

\subsection{Consensus with Constraints}

We examine the following consensus problem with constraints where $\mathcal{X}=[-50,50]$, i.e., here $\eta=50$. The legitimate agents aim to minimize the function 
\[\min_{x\in[-50,50]}\frac{1}{|\LL|}\sum_{i\in\LL}f_i(x),\]  where  $f_i(x)=\frac{1}{2}(x-b_i)^2$
and
$b_i=\tilde{b}_i$ for all $i\in\LL$.

The optimal nominal value is
$x_{\LL}^{\star} = x_{\LL}^{\star\text{UC}}\approx 31.367$ for our choice of $(b_i)_{i=1}^{15}$,  where we round the solution to the second digit after the decimal point.
Consequently,  the dynamic \eqref{eq:legitimate_update_opt_gradient_selfish_trusted} can be written as follows
    \begin{flalign}\label{eq:update_1d_numrical}
    c_{i}(t)&= \sum_{j\in\NN_i(t)\cup \{i\}}\hspace{-0.45cm}w_{ij}(t)x_j(t)\nonumber\\
    y_{i}(t)&=c_i(t)-\frac{2\cdot\mathds{1}_{\{t-T_0 \geq 0\}}}{(1-\epsilon)(t-T_0+2)}\cdot(c_i(t)-b_i),\nonumber\\
    x_i(t+1) &= \mathds{1}_{\{y_{i}(t)\in\XX\}}y_{i}(t)+\text{sign}(y_{i}(t))\eta\mathds{1}_{\{y_{i}(t)\notin\XX\}}
    \end{flalign}
for the choice $\gamma(t)=\frac{2}{\mu_{\epsilon}(t+2)}=\frac{2}{\mu(1-\epsilon)(t+2)}$ where $\mu=1, \epsilon=0.1$.    
The initial values of the legitimate agents are chosen randomly and uniformly in the interval $[-\eta,\eta]$.
Note that in this setup the optimal point lies in the constraint set $\XX$. Nonetheless, the set $\XX$ affects the update rule \eqref{eq:update_1d_numrical} and limits the data values at all times to be in the set $\XX$.

To maximize the harmful impact of malicious agents on our analytical results, we choose the malicious agents' values to be equal to $-50$, i.e., $-\eta$, at all times. We choose this easy-to-spot malicious attack since it maximizes the deviation of the malicious inputs from the optimal nominal solution. 
Nonetheless, our algorithm can tolerate \emph{arbitrary} malicious node inputs including the time-varying case or small deviation case that is usually much harder to detect. 

As a benchmark, we compare the performance of our proposed Algo.~\ref{alg:agent_i_dynamic} to that of \cite{Sundaram_distributed_opt_malicious_2019} which adapts the  W-MSR consensus algorithm \cite{leBlancWMSR} to the case of one-dimensional distributed optimization problems where $d=1$.

\subsubsection{Trust Values with Uniform Distribution}
First, we consider the setup in the main paper where the trust observations are distributed as follows. Let $\EEop[\alpha_{ij}(t)] = 0.55$ if $j\in\mathcal{N}_i\cap \LL$, 
and $\EEop[\alpha_{ij}(t)] = 0.45$ if $j\in\mathcal{N}_i\cap \MM$. The random variable $\alpha_{ij}(t)$ is uniformly distributed on the interval $\left[\EEop[\alpha_{ij}(t)]-\frac{\ell}{2},\EEop[\alpha_{ij}(t)]+\frac{\ell}{2}\right]$, for every $i\in\LL$ and $j\in\mathcal{N}_i$. We consider  the values $\ell=0.6$ and $\ell=0.8$, in both scenarios $|E_{\LL}|=|E_{\MM}|=0.05$.

\begin{figure}[t!]
    \centering
    \includegraphics[scale=0.6,trim={0.05cm 0cm 0cm 0.03cm},clip]{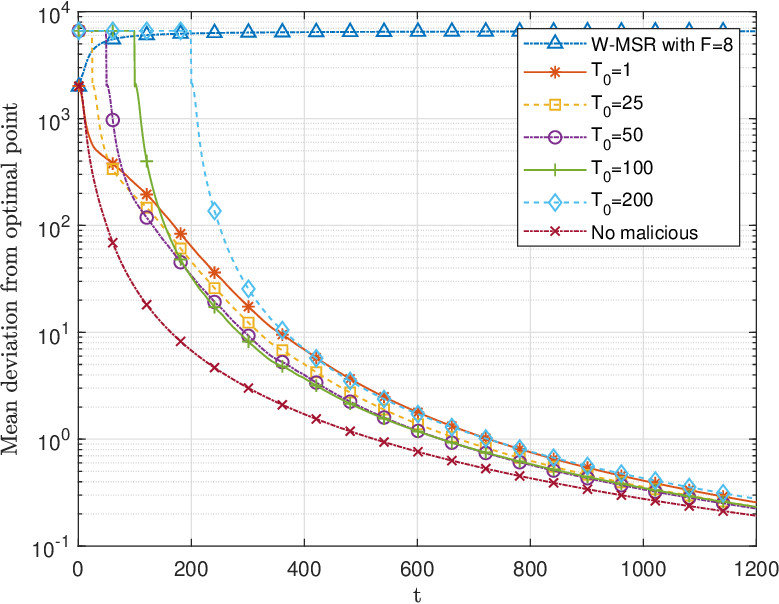}
    \caption{{\small Average  $\overline{e}(t)$ of as a function of $t$ for $|\MM|=15$ and the noise in the stochastic trust value is chosen as $\ell=0.8$.}}
    \label{fig:fig_15_malicious}
\end{figure}

\begin{figure}[t!]
    \centering
    \includegraphics[scale=0.6,trim={0.05cm 0cm 0cm 0.03cm},clip]{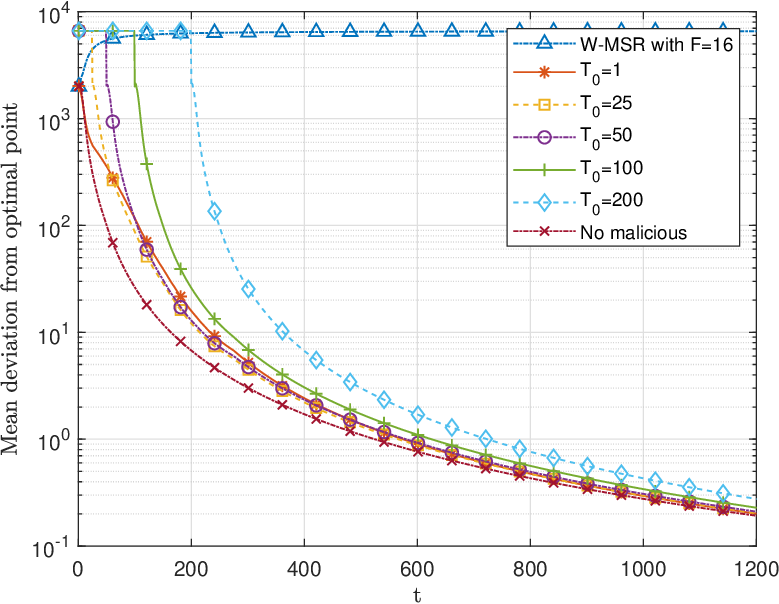}
    \caption{{\small Average  $\overline{e}(t)$ as a function of $t$ for $|\MM|=30,$ and the noise in the stochastic trust value is chosen as $\:\ell=0.6$.}}
    \label{fig:fig_30_malicious}
    %\vspace{-0.7cm}
\end{figure}

Figs.~\ref{fig:fig_15_malicious} and \ref{fig:fig_30_malicious} capture the average mean squared distance of the legitimate agents from the optimal point $x_{\LL}^{\star}$ for each time $t$.
We can see that the W-MSR algorithm fails to converge to the optimal solution (i.e., nominal consensus point). This occurs due to the high number of malicious agents, which is higher in this case than the tolerance threshold\footnote{We can upper bound the tolerance threshold for this setup by $2$ following our argument in \cite{ourTRO}.} in \cite{Sundaram_distributed_opt_malicious_2019}. Additionally, generally the W-MSR algorithm is not guaranteed to converge to the optimal value $x_{\LL}^{\star}$ but to a value in the convex hull of $\Pi_{[-\eta,\eta]}(b_i),\: i\in\LL$. 
In our case, this convex hull is exactly the interval $[-\eta,\eta]$, thus the W-MSR algorithm cannot guarantee the reduction of the distance to the optimal value with respect to the interval $[-\eta,\eta]$. In contrast, Algo.~\ref{alg:agent_i_dynamic} provides resilience to malicious activity and can tolerate even $30=2|\LL|$ malicious agents, as evident  in Figs.~\ref{fig:fig_15_malicious} and \ref{fig:fig_30_malicious}. Furthermore, we can see from Figs.~\ref{fig:fig_15_malicious} and \ref{fig:fig_30_malicious} that Algo.~\ref{alg:agent_i_dynamic} is robust to small values of $|E_{\LL}|$ and $|E_{\MM}|$. Finally, Figs.~\ref{fig:fig_15_malicious} and \ref{fig:fig_30_malicious} show the effect of the variance of the trust values on the incentive to increase $T_0$. In Fig.~\ref{fig:fig_15_malicious}, we can benefit from increasing $T_0$ up to $100$.
This occurs since the higher variance of the trust values increases the misclassification errors. 
Since the probability of these errors decreases with $T_0$, they are less impactful when $T_0$ is  $100$ or higher.
However, we can see that when we decrease in Fig.~\ref{fig:fig_30_malicious} the variance of the trust values by setting $\ell=0.6$, it is not beneficial to increase $T_0$ beyond $25$.
Note that regardless of the value of the observation window $T_0$, our algorithm eventually recovers the global optimum as predicted by theory. Thus, the value of $T_0$ mostly dictates \emph{the rate} of recovery of the global optimum.

\subsubsection{Trust Values with Bernoulli Distribution}

\begin{figure}[t!]
    \centering
    \includegraphics[scale=0.6,trim={0.05cm 0cm 0cm 0.03cm},clip]{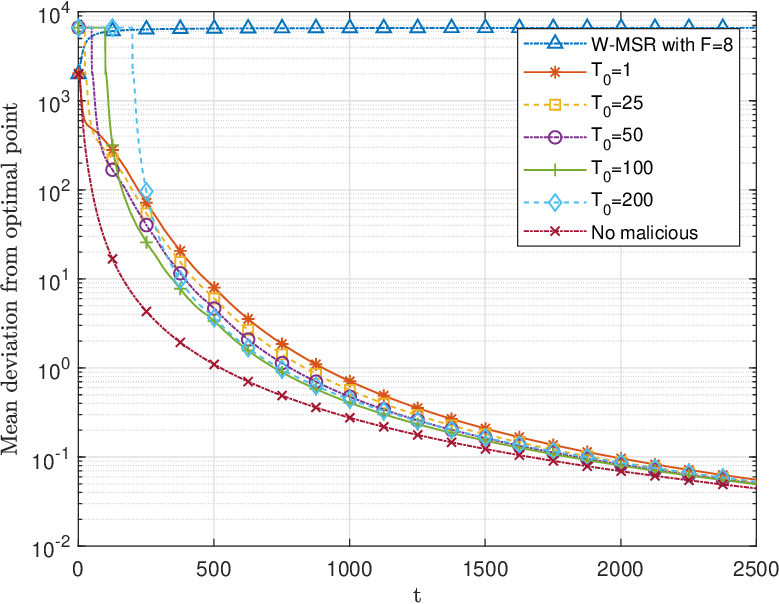}
    \caption{{\small Average  $\overline{e}(t)$ of as a function of $t$ for $|\MM|=15$ Bernoulli trust observations.}}
    \label{fig:fig_15_malicious_bernoulli}
    \end{figure}

\begin{figure}[t!]
    \centering
    \includegraphics[scale=0.6,trim={0.05cm 0cm 0cm 0.03cm},clip]{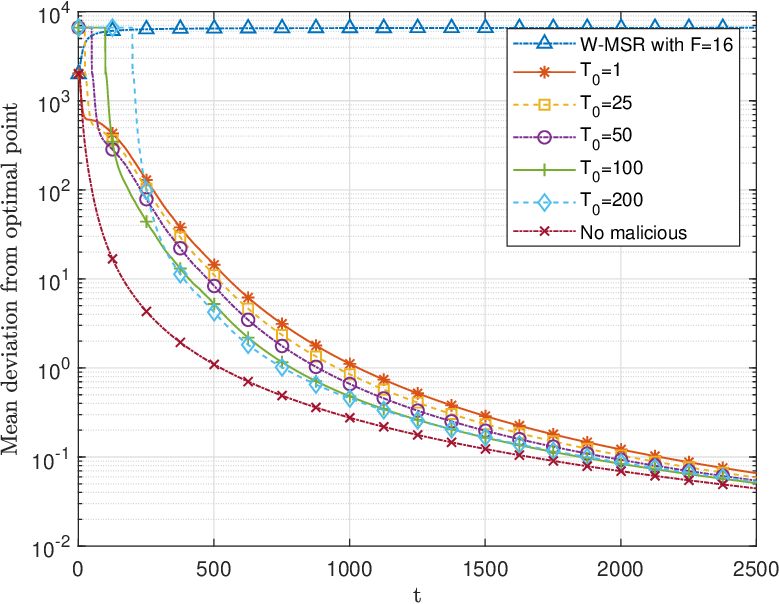}
    \caption{{\small Average  $\overline{e}(t)$ as a function of $t$ for $|\MM|=30$ for Bernoulli trust observations.}}
    \label{fig:fig_30_malicious_bernoulli}
    %\vspace{-0.7cm}
\end{figure}

We now consider trust value observations that are distributed according to Bernoulli random variables. That is, $\alpha_{ij}(t)\sim \text{Bernoulli}(p_{ij})$ where $p_{ij}=0.6$ for $i,j\in\LL$, i.e., both $i$ and $j$ are legitimate, and $p_{ij}=0.4$ for $i\in\LL$ and $j\in\MM$.
The importance and uniqueness of this choice of distribution stem from the fact that the Bernoulli distribution with parameter $p$ has the highest variance, i.e., $p(1-p)$, among all the distributions defined over the sample space $[0,1]$ with expectation $p$. Overall, this setup has larger variances than the one we examined in the previous subsection, even when taking into account the slightly improved values of $|E_{\LL}|=|E_{\MM}|=0.1$ it leads to. Figs.~\ref{fig:fig_15_malicious_bernoulli} and \ref{fig:fig_30_malicious_bernoulli} show that in this setup, increasing $T_0$ even beyond $100$ can reduce the average mean squared error as opposed to Figs.~\ref{fig:fig_15_malicious} and \ref{fig:fig_30_malicious} where increasing $T_0$ could lead to increased average mean squared error evaluations which are noticeable for smaller running time $t$.

\subsection{Evaluating Analytical Bounds}

Generally, upper bounds such as \eqref{eq:upper_bound_deviation_no_malicious}, \eqref{eq:upper_bound_suboptimality_gap_cor1}--\eqref{eq:upper_bound_suboptimality_gap_cor3}, and \eqref{eq:upper_bound_suboptimality_gap_alternative} are utilized to understand the rate of convergence and are not necessarily tight in a finite time regime. In particular, their dependence on the worst-case scenario in using constants such as the maximal gradient and the second largest eigenvalue modulus of the weight matrix leads to somewhat loose bounds for short-time regimes, even in the nominal case where no malicious agents are present. In what follows we examine the relations between the upper bounds \eqref{eq:upper_bound_deviation_no_malicious}, \eqref{eq:upper_bound_suboptimality_gap_cor1}--\eqref{eq:upper_bound_suboptimality_gap_cor3}, and \eqref{eq:upper_bound_suboptimality_gap_alternative}. We will see that these bounds exhibit a more pessimistic view of the average mean squared error when compared with the numerical evaluation of Algorithm \ref{alg:agent_i_dynamic}. Nonetheless, the intuitions they lead to are valuable in understanding and forecasting how different system components, such as the precision of the trust observation, the connectivity, etc., influence it. 
For our comparison, we chose the simple one-dimensional setup we presented in the previous subsection, where $|E_{\LL}|$ and $|E_{\MM}|$ are set to $0.1$. Thus, $\EEop(\alpha_{ij}(t))=0.6$ for $i,j\in\LL$, and $\EEop(\alpha_{ij}(t))=0.4$ for $i\in\LL$ and $j\in\MM$. For this setup, we have that $G=193.3$, $\eta=50$ and $\rho_{\LL}\approx 0.9$, and $\mu=L=1$. We note that since $G^2\approx 3.7\times 10^4$ and $(1-\rho)^4\approx 10^{-4}$ the dominant constant in the nominal case is $ 2\cdot\frac{\mu(1/\epsilon-1)+L}{(1-\epsilon)^2}\cdot\frac{G^2}{\mu^2(1-\rho_{\LL})^4}\approx 9.1\times 10^9$, and thus taking the other constant into account, we will need approximately $10^6$ iterations until the effect of the constants, i.e., the terms that do not depend on the number of iterations $T$, ceases to dominate the upper bound \eqref{eq:upper_bound_deviation_no_malicious}.

 \paragraph{$O(1/T)$ expected convergence rate}
 
\begin{figure}[h]
  \begin{subfigure}{.5\textwidth}
  \centering
    \includegraphics[width=.9\linewidth]{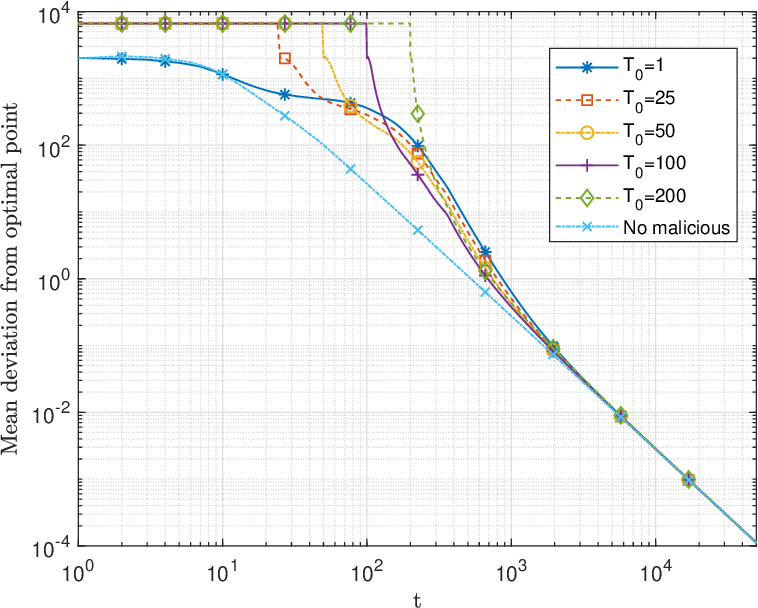}
    \caption{Numerical: average  $\overline{e}(t)$ as a function of $t$.}
    \label{fig:log_fig_15_malicious_bernoulli}
  \end{subfigure}%
  \begin{subfigure}{.5\textwidth}
  \centering
    \includegraphics[width=.9\linewidth]{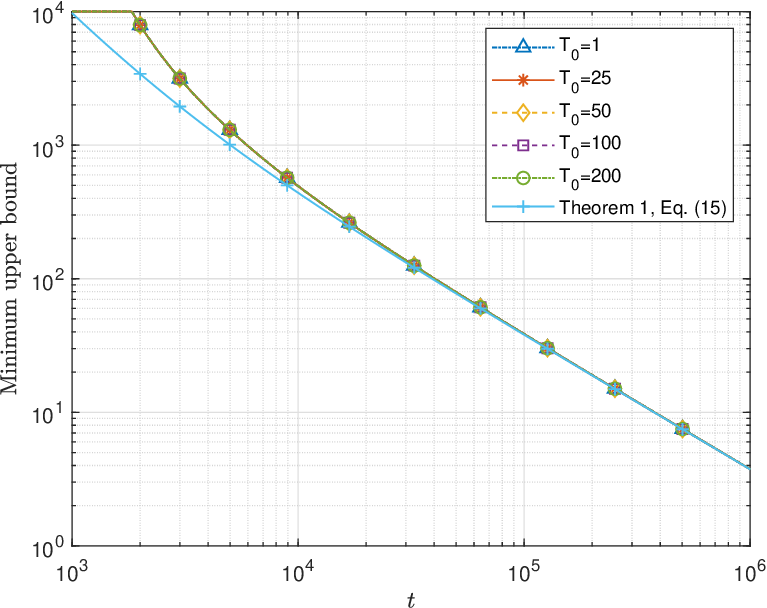}
    \caption{The minimum analytical upper bound of \eqref{eq:upper_bound_suboptimality_gap_cor1}--\eqref{eq:upper_bound_suboptimality_gap_cor3}, and \eqref{eq:upper_bound_suboptimality_gap_alternative}.}
    \label{fig:log_fig_15_malicious_analytical}
  \end{subfigure}
  \caption{Convergence rate: a system with 15 malicious agents}
  \label{fig:convergence_rate_comparison_15_malicious}
\end{figure}

\begin{figure}[h]
  \begin{subfigure}{.5\textwidth}
  \centering
    \includegraphics[width=.9\linewidth]{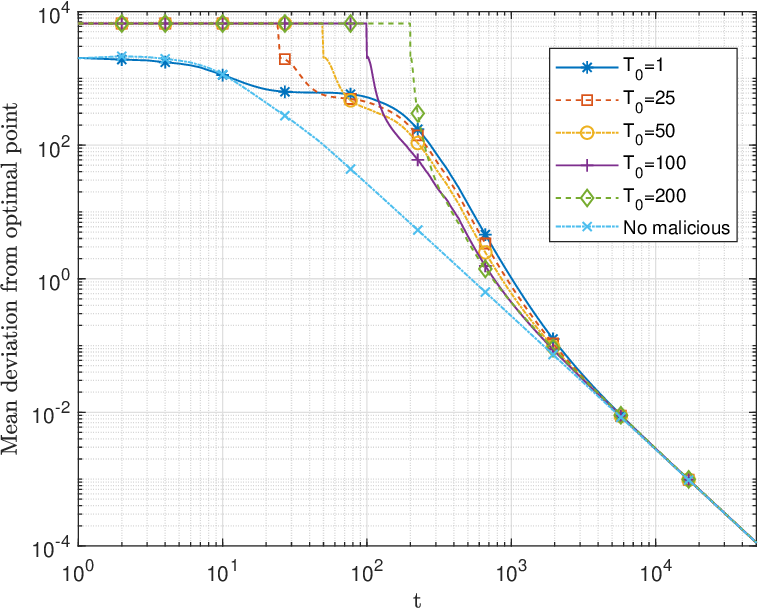}
    \caption{Numerical: average  $\overline{e}(t)$ as a function of $t$.}
    \label{fig:log_fig_30_malicious_bernoulli}
  \end{subfigure}%
  \begin{subfigure}{.5\textwidth}
  \centering
    \includegraphics[width=.9\linewidth]{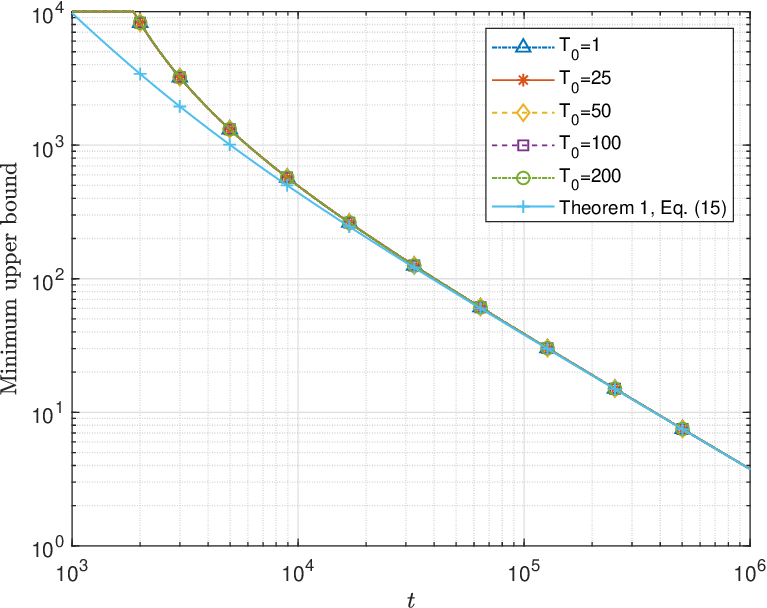}
    \caption{The minimum analytical upper bound of \eqref{eq:upper_bound_suboptimality_gap_cor1}--\eqref{eq:upper_bound_suboptimality_gap_cor3}, and \eqref{eq:upper_bound_suboptimality_gap_alternative}.
    \label{fig:log_fig_30_malicious_analytical}
    }
  \end{subfigure}
  \caption{Convergence rate: A system with 30 malicious agents}
    \label{fig:convergence_rate_comparison_30_malicious}
\end{figure}

We first examine the convergence rate analysis and confirm its linearity.  To this end,  in Figs.\ref{fig:log_fig_15_malicious_bernoulli} and 
\ref{fig:log_fig_30_malicious_bernoulli} we re-plot
Figs.~\ref{fig:fig_15_malicious_bernoulli} and \ref{fig:fig_30_malicious_bernoulli}, respectively, in a logarithmic scale of both axes (and omit the W-MSR plot which is redundant for this comparison). We compare these numerical plots to the analytical bounds by plotting in Figs. \ref{fig:log_fig_15_malicious_analytical} and \ref{fig:log_fig_30_malicious_analytical} at each time the minimal upper bound among  \eqref{eq:upper_bound_suboptimality_gap_cor1}--\eqref{eq:upper_bound_suboptimality_gap_cor3}, and \eqref{eq:upper_bound_suboptimality_gap_alternative}. As a baseline, we additionally plot in Figs.\ref{fig:log_fig_15_malicious_analytical} and \ref{fig:log_fig_30_malicious_analytical} the analytical upper bound \eqref{eq:upper_bound_deviation_no_malicious} for a system with no malicious agents. Figs. \ref{fig:convergence_rate_comparison_15_malicious} and \ref{fig:convergence_rate_comparison_30_malicious} validate both numerically and analytically the $O(1/T)$ expected convergence rate.

\paragraph{Increased inter-connectivity of legitimate agents reduces the average mean square distance to the optimal point}
Lastly, we examine the effect of increasing the connectivity among the legitimate agents both numerically, and analytically (according to the upper bounds \eqref{eq:upper_bound_suboptimality_gap_cor1}--\eqref{eq:upper_bound_suboptimality_gap_cor3} and~\eqref{eq:upper_bound_suboptimality_gap_alternative}). 
To this end,  we consider the topology we have considered so far that is captured in Fig.~\ref{fig:connectivity_graph} where $\rho_{\LL}\approx 0.89$ where some legitimate agents are connected to others, and every legitimate agent is connected to all malicious agents. Additionally, we increase the connectivity among the legitimate agents and consider
a fully connected connectivity graph where every legitimate agent is connected to all other agents (legitimate or malicious); in this scenario $\rho_{\LL}=0.5$.

\begin{figure}[h]
  \begin{subfigure}{.5\textwidth}
  \centering
    \includegraphics[width=.9\linewidth]{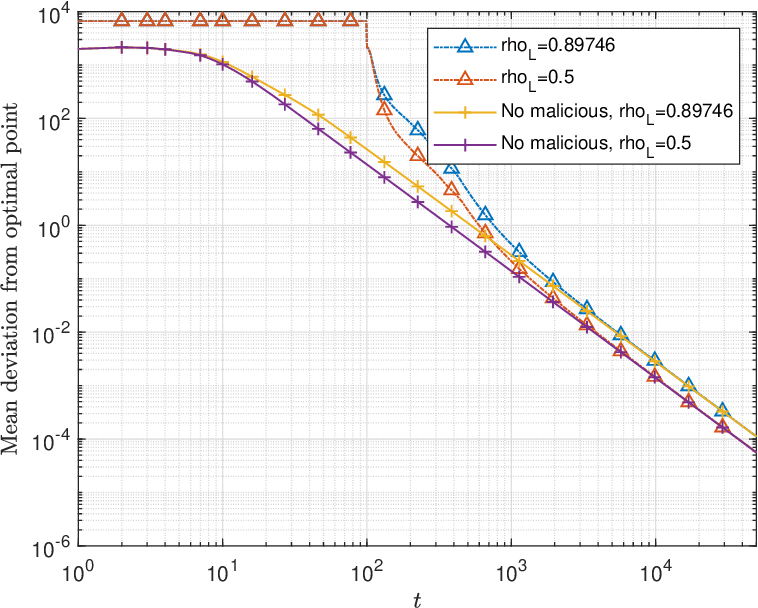}
    \caption{Numerical: average  $\overline{e}(t)$ as a function of $t$.}
\label{fig:log_fig_15_malicious_bernoulli_connectivity}
  \end{subfigure}%
  \begin{subfigure}{.5\textwidth}
  \centering
    \includegraphics[width=.9\linewidth]{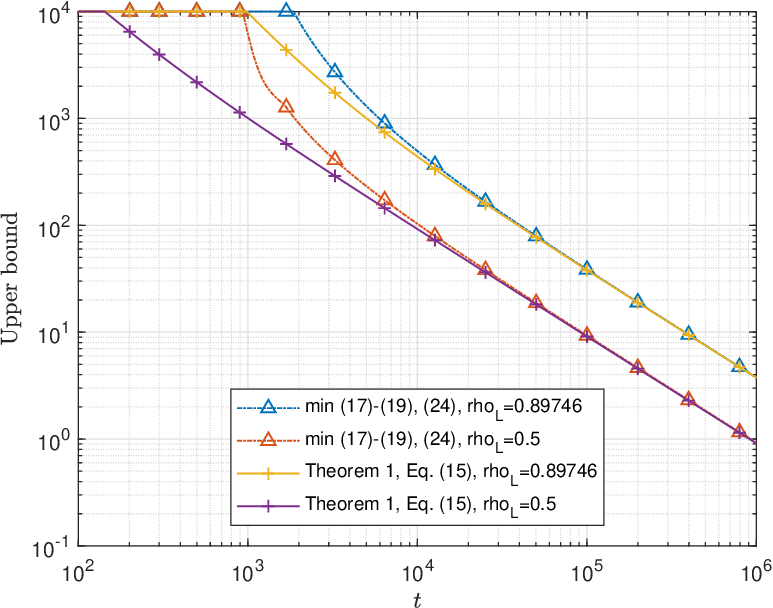}
    \caption{The minimum analytical upper bound of \eqref{eq:upper_bound_suboptimality_gap_cor1}--\eqref{eq:upper_bound_suboptimality_gap_cor3}, and \eqref{eq:upper_bound_suboptimality_gap_alternative}.}
\label{fig:log_fig_15_malicious_analytical_connectivity}
  \end{subfigure}
  \caption{A system with 15 malicious agents and $T_0=100$}
\label{fig:convergence_rate_comparison_15_malicious_connectivity}
\end{figure}

\begin{figure}[h]
  \begin{subfigure}{.5\textwidth}
  \centering
    \includegraphics[width=.9\linewidth]{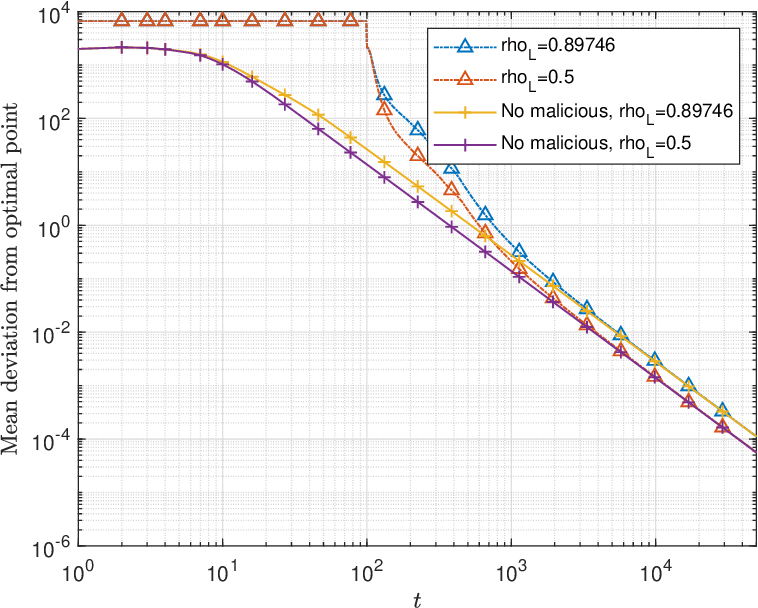}
    \caption{Numerical: average  $\overline{e}(t)$ as a function of $t$.}
\label{fig:log_fig_30_malicious_bernoulli_connectivity}
  \end{subfigure}%
  \begin{subfigure}{.5\textwidth}
  \centering
    \includegraphics[width=.9\linewidth]{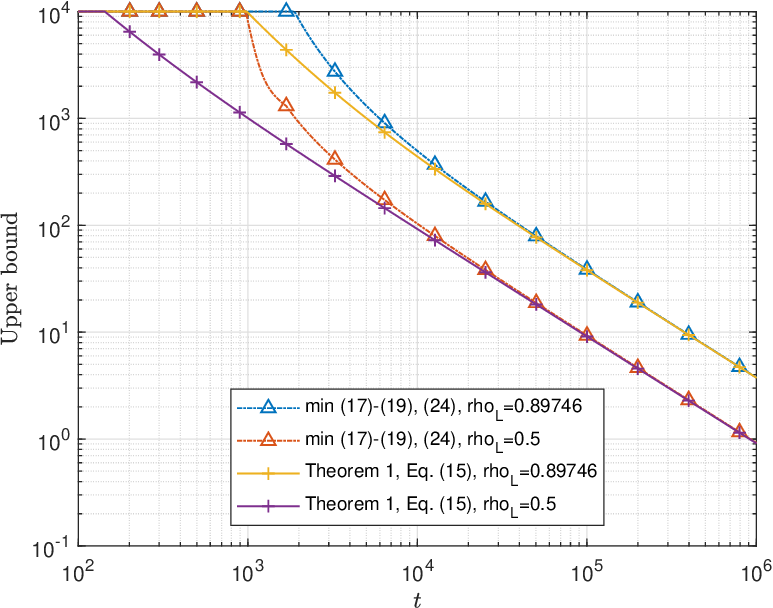}
    \caption{The minimum analytical upper bound of \eqref{eq:upper_bound_suboptimality_gap_cor1}--\eqref{eq:upper_bound_suboptimality_gap_cor3}, and \eqref{eq:upper_bound_suboptimality_gap_alternative}.
\label{fig:log_fig_30_malicious_analytical_connectivity}
    }
  \end{subfigure}
  \caption{A system with 30 malicious agents and $T_0=100$}
\label{fig:convergence_rate_comparison_30_malicious_connectivity}
\end{figure}

Figs.~\ref{fig:convergence_rate_comparison_15_malicious_connectivity} and~\ref{fig:convergence_rate_comparison_30_malicious_connectivity} illustrate how increasing the connectivity between the legitimate agents reduces the mean squared error (in a logarithmic scale); for the sake of clarity of presentation, we present the results for $T_0=100$. Specifically, Figs.~\ref{fig:log_fig_15_malicious_bernoulli_connectivity} and~\ref{fig:log_fig_30_malicious_bernoulli_connectivity} capture the influence of the increased connectivity of the legitimate agents on the mean squared error of Algo.~\ref{alg:agent_i_dynamic} numerically for systems with $15$ and $30$ malicious agents, respectively. Similarly, Figs.~\ref{fig:log_fig_15_malicious_analytical_connectivity} and~\ref{fig:log_fig_30_malicious_analytical_connectivity} present the effect of increasing the connectivity of the legitimate agents on the minimum analytical upper bound of \eqref{eq:upper_bound_suboptimality_gap_cor1}--\eqref{eq:upper_bound_suboptimality_gap_cor3} and~\eqref{eq:upper_bound_suboptimality_gap_alternative}, for systems with $15$ and $30$ legitimate agents, respectively. We can observe from Figs. \ref{fig:convergence_rate_comparison_15_malicious_connectivity} and \ref{fig:convergence_rate_comparison_30_malicious_connectivity} that increasing the connectivity among the legitimate agents decreases the mean squared error in both the numerical evaluation and analytical bounds. Thus, the intuition that the analytical bounds provide also converts to the numerical evaluation of Algo.~\ref{alg:agent_i_dynamic}. We note, however, that the anticipated improvement in performance exhibited in Figs.~\ref{fig:log_fig_15_malicious_analytical_connectivity} and~\ref{fig:log_fig_30_malicious_analytical_connectivity} is more modest in reality, see Figs.~\ref{fig:log_fig_15_malicious_bernoulli_connectivity} and~\ref{fig:log_fig_30_malicious_bernoulli_connectivity}. We believe that this occurs due to the pessimism of many upper bounds in the literature including \eqref{eq:upper_bound_suboptimality_gap_cor1}--\eqref{eq:upper_bound_suboptimality_gap_cor3} and \eqref{eq:upper_bound_suboptimality_gap_alternative}, where the connectivity is captured by multiplication of terms such $(1-\rho_{\LL})^{k}, \: k\in\mathbb{N}$ that is not necessarily tight; additionally, this multiplication also amplify other pessimistic terms in \eqref{eq:upper_bound_suboptimality_gap_cor1}--\eqref{eq:upper_bound_suboptimality_gap_cor3} and~\eqref{eq:upper_bound_suboptimality_gap_alternative}, such as $E_{\LL},E_{\MM}$ and $G$.

% \vspace{-0.5cm}

% \section*{References}
% \vspace{-0.7cm}

% \bibliographystyle{IEEEtran}
% %\bibliographystyle{alpha}
% \bibliography{references}

%% file: commands.tex
\newcommand\FF{\mathcal{F}}
\newcommand\CC{\mathcal{C}}
\newcommand\PP{\mathcal{P}}
\newcommand\UU{\mathcal{U}}
\newcommand\WW{\mathcal{W}}
\newcommand\MM{\mathcal{M}}
\newcommand\NN{\mathcal{N}}
\newcommand\DD{\mathcal{D}}
\newcommand\QQ{\mathcal{Q}}
\newcommand\XX{\mathcal{X}}
\newcommand\YY{\mathcal{Y}}
\newcommand\ZZ{\mathcal{Z}}
\newcommand\GG{\mathcal{G}}
\newcommand\II{\mathcal{I}}
\newcommand\LL{\mathcal{L}}
\newcommand\BB{\mathcal{B}}
\newcommand\VV{\mathcal{V}}
\newcommand\EE{\mathcal{E}}
\newcommand{\RR}{\mathcal{R}}
\newcommand{\HH}{\mathcal{H}}
\renewcommand\SS{\mathcal{S}}
\newcommand\EEop{\mathbf{E}}

\newcommand\TCB[1]{\textcolor{blue}{#1}}

%% Notation macros %%%
\def\limsup{\mathop{\rm limsup}}
\def\1{\mathbf{1}}
\newcommand{\nm}{n_{\MM}}
\newcommand{\G}{\mathbb{G}}
\newcommand{\A}{\mathcal{A}}
\newcommand{\V}{\mathbb{V}}
\newcommand{\W}{W}
\newcommand{\wij}{w_{ij}}
\newcommand{\Ni}{\mathcal{N}_i}
\newcommand{\xL}{x_\mathcal{L}}
\newcommand{\xM}{x_\MM}
\newcommand{\WL}{W_\mathcal{L}}
\newcommand{\WM}{W_\MM}
\newcommand{\aij}{\alpha_{ij}}
\newcommand{\dev}{\Delta(T_0,\delta)}
\newcommand{\maxdev}{\Delta_\text{max}(T_0,\delta)}

\def\argmin{{\mathop {\rm argmin}}}

\newcommand{\norm}[1]{\left\lVert#1\right\rVert}

%% file: append_classification_prob.tex
\section{}\label{append:proof_lemma_classification_prob}

\begin{proof}[Proof of Lemma \ref{lemma:classification_prob}]
Denote for every $k\geq0$,
\[\EE(k)\triangleq\bigcup_{\substack{i\in\LL,\\j\in\mathcal{N}_i\cap\LL}}\{\beta_{ij}(k)<0\}
\bigcup_{\substack{i\in\LL,\\j\in\mathcal{N}_i\cap\MM}}\{\beta_{ij}(k)\geq0\}.\]
By Lemma \ref{Lemma:concentration_upper}, for all $k\geq1$ we have that 
\begin{flalign}\label{eq:prob_miscalssifying_T_f_proof}
&\Pr(T_f=k) \stackrel{(a)}{\leq}  \Pr\Bigg(\EE(k-1)\Bigg)\nonumber\\
&\stackrel{(b)}{\leq} \sum_{\substack{i\in\LL,\\j\in\mathcal{N}_i\cap\LL}}\hspace{-0.1cm}\Pr(\beta_{ij}(k-1)<0)+\hspace{-0.1cm}\sum_{\substack{i\in\LL,\\j\in\mathcal{N}_i\cap\MM}}\hspace{-0.1cm}\Pr(\beta_{ij}(k-1)\geq0)\nonumber\\
&\stackrel{(c)}{\leq}\sum_{i\in\LL}|\mathcal{N}_i\cap\LL|\exp(-2(k-1)E_{\LL}^2)\nonumber\\
&\qquad+\sum_{i\in\LL}|\mathcal{N}_i\cap\MM|\exp(-2(k-1)E_{\MM}^2),
\end{flalign}
where $(a)$ follows from the definition of $T_f$ in Corollary \ref{cor:rand_finite_classifications} which implies that if $T_f=k$ there must be a misclassification error in the legitimacy of agents at time $k-1$,  $(b)$ follows from the union bound, and (c) follows from Lemma \ref{Lemma:concentration_upper}. 

Now, since by definition, $p_c(-1)\geq1$ we can conclude \eqref{eq:prob_miscalssifying_T_f}.

From the definition of $T_f$ in Corollary \ref{cor:rand_finite_classifications} $\Pr(T_f>k-1)=1$ for $k=0$. Additionally, by the union bound, for all $k\geq 1$,
\begin{flalign}\label{eq:T_f_greater_t_proof}
&\Pr(T_f>k-1) = \Pr\left(\bigcup_{t\geq k-1} \EE(t)\right)\leq \sum_{t=k-1}^{\infty}\Pr\left( \EE(t)\right)\nonumber\\
&\leq\sum_{t=k-1}^{\infty}\:\sum_{i\in\LL}|\mathcal{N}_i\cap\LL|\exp(-2tE_{\LL}^2)\nonumber\\
&\qquad+\sum_{t=k-1}^{\infty}\:\sum_{i\in\LL}|\mathcal{N}_i\cap\MM|\exp(-2tE_{\MM}^2)\nonumber\\
&=D_{\LL}\frac{\exp(-2(k-1)E_{\LL}^2)}{1-\exp(-2E_{\LL}^2)}+D_{\MM}\frac{\exp(-2(k-1)E_{\MM}^2)}{1-\exp(-2E_{\MM}^2)}.
\end{flalign}
Observe that $\Pr(T_f>k-1)$ vanishes as $k$ tends to infinity.
\end{proof}

%% file: append_convergence_in_mean_proof_definition.tex
\section{}\label{proof:converegence_mean_by_definition}
\begin{proof}[Proof by Definition of Theorem \ref{theorem:convergence_mean}]
Let us assume that $t\geq 2T_0$. Denote $M_k\triangleq\max\{k,T_0\}$ and  $t/2\triangleq\lfloor\frac{t}{2}\rfloor$. 
Next, we utilize the law of total expectation as follows:
\begin{flalign*}
&\EEop\left[\|x_{i}(t)-x_{\LL}^{\star}\|^r\right]=\EEop\left[\EEop[\|x_{i}(t)-x_{\LL}^{\star}\|^r\mid T_f]\right]\nonumber\\
&= \sum_{k=0}^{t-1}\Pr(T_f=k)\EEop\left[\|x_{i}(t)-x_{\LL}^{\star}\|^r\:\big| \:T_f=k\right]\nonumber\\
&\quad+ \Pr(T_f>t-1)\EEop\left[\|x_{i}(t)-x_{\LL}^{\star}\|^r\:\big| \:T_f>t-1\right]\nonumber\\
&= \sum_{k=0}^{t-1}\Pr(T_f=k)\EEop\left[\|z_i(t-M_k)-x_{\LL}^{\star}\|^r\:\big| \:T_f=k\right]\nonumber\\
&\quad+ \Pr(T_f>t-1)\EEop\left(\|x_{i}(t)-x_{\LL}^{\star}\|^r\:\big| \:T_f>t-1\right)\nonumber\\
&\leq \sum_{k=0}^{t-1}\Pr(T_f=k)\EEop\left[\|z_i(t-M_k)-x_{\LL}^{\star}\|^r\:\big| \:T_f=k\right]\nonumber\\
&\quad+ \Pr(T_f>t-1)\left(2\eta\right)^r,
\end{flalign*}
where $z_i(0)=x_{i}(M_k)$ for $i\in\LL$.

Next, we upper bound the term $\sum_{k=0}^{t-1}\Pr(T_f=k)\EEop\left[\|z_i(t-M_k)-x_{\LL}^{\star}\|^r\:\big| \:T_f=k\right]$ utilizing the upper bounds \eqref{eq:prob_miscalssifying_T_f} and \eqref{eq:upper_bound_deviation_no_malicious} which hold for every initial point $z_i(0)\in\XX$:
\begin{flalign*}
&\sum_{k=0}^{t-1}\Pr(T_f=k)\EEop\left[\|z_i(t-M_k)-x_{\LL}^{\star}\|^r\:\big| \:T_f=k\right]\nonumber\\
\leq&\sum_{k=0}^{t/2}\Pr(T_f=k)\left[\frac{4\overline{h}_{\epsilon}(t-M_k)|\LL|}{\mu_{\epsilon}(t-M_k)(t-M_k+1)}\right]^{\frac{r}{2}}\nonumber\\
&+\sum_{k=t/2+1}^{t-1}\Pr(T_f=k)\cdot\min\left\{\left[\frac{4\overline{h}_{\epsilon}(t-M_k)|\LL|}{\mu_{\epsilon}(t-M_k)(t-M_k+1)}\right]^{\frac{r}{2}},(2\eta)^r\right\}.
\end{flalign*}
Now, since $t\geq 2T_0$, $\mu_{\epsilon}>0$ and $\frac{\overline{h}_{\epsilon}(t)}{t(t+1)}$ is a decreasing function we have that
\begin{flalign*}
&\sum_{k=0}^{t/2}\Pr(T_f=k)\left[\frac{4\overline{h}_{\epsilon}(t-M_k)|\LL|}{\mu_{\epsilon}(t-M_k)(t-M_k+1)}\right]^{\frac{r}{2}}\nonumber\\
&\leq \left[\frac{4\overline{h}_{\epsilon}(t/2)|\LL|}{\mu_{\epsilon}(t/2)(t/2+1)}\right]^{\frac{r}{2}}\sum_{k=0}^{t/2}\Pr(T_f=k)\leq \left[\frac{4\overline{h}_{\epsilon}(t/2)|\LL|}{\mu_{\epsilon}(t/2)(t/2+1)}\right]^{\frac{r}{2}}.
\end{flalign*}
Furthermore,
\begin{flalign*}
&\sum_{k=t/2+1}^{t-1}\Pr(T_f=k)\cdot\min\left\{\left[\frac{4\overline{h}_{\epsilon}(t-M_k)|\LL|}{\mu_{\epsilon}(t-M_k)(t-M_k+1)}\right]^{\frac{r}{2}},(2\eta)^r\right\}\nonumber\\
&\leq(2\eta)^r\sum_{k=t/2+1}^{t-1} \Pr(T_f=k)\leq(2\eta)^r\Pr(T_f>t/2)\leq(2\eta)^r p_e(t/2).
\end{flalign*}
Additionally by \eqref{eq:T_f_greater_t}, $
\Pr(T_f>t-1)\left[2\eta\right]^r\leq \left[2\eta\right]^r p_e(t-1)$.
Consequently, 
$\lim_{t\rightarrow\infty}\EEop\left[\|x_{i}(t)-x_{\LL}^{\star}\|^r\right]=0$, $\forall\: r\geq1$.
\end{proof}

%% file: append_distance_mean.tex
\section{}\label{append:ditance_t_mean_proof}

\begin{proof}[Proof of Proposition \ref{prop:distance_to_mean_time_t}]
First, we utilize the identity
\begin{flalign}\label{eq:upper_L21_frobenius2}
\sum_{j\in\LL}\|[A]_j\|^2=\|A\|^2_{\text{F}},
\end{flalign}
and the upper bound $\frac{1}{2}(\|[A]_i\|^2+\|[A]_j\|^2)\geq\|[A]_i\|\cdot \|[A]_j\|$ to deduce that
\begin{flalign}\label{eq:upper_L21_frobenius1}
\sum_{j\in\LL}\|[A]_j\|\leq \sqrt{|\LL|}\cdot\|A\|_{\text{F}}.
\end{flalign}
Thus, in what follows, we focus our efforts on upper bounding the  Frobenius norms $\|X(t)-\overline{X}(t)\|_{\text{F}}$ and $\|X(t)-\overline{X}(t)\|_{\text{F}}^2$.

Observe that 
\begin{flalign}\label{eq:delta_t_frob_upper}
\left\|\Delta(t)-\frac{\Delta(t)\boldsymbol{1}}{|\LL|}\boldsymbol{1}^T\right\|_{\text{F}}&=\sqrt{\sum_{j\in\LL}\left\|[\Delta(t)]_j-\frac{1}{|\LL|}\sum_{k\in\LL}[\Delta(t)]_k\right\|^2}\nonumber\\
&\leq \sqrt{\sum_{j\in\LL}\left\|[\Delta(t)]_j\right\|^2}=\left\|\Delta(t)\right\|_{\text{F}}.
\end{flalign}
Additionally,  by the singular value decomposition of the doubly-stochastic matrix $W(t)$ we have that 
\begin{flalign}\label{eq:upper_weight_prod_sterp_rho_upper}
&\left\|\left(X(t)-\frac{X(t)\boldsymbol{1}}{|\LL|}\boldsymbol{1}^T\right)W^T(t)\right\|_{\text{F}}\nonumber\\
&= \left\|W(t)\left(X(t)-\frac{X(t)\boldsymbol{1}}{|\LL|}\boldsymbol{1}^T\right)^{T}\right\|_{\text{F}}\nonumber\\
&\leq \sqrt{\sum_{i=1}^d\left\|W(t)\left([X^T(t)]_i-\boldsymbol{1}\frac{\boldsymbol{1}^T[X^T(t)]_i}{|\LL|}\right)\right\|^2}\nonumber\\
&\leq \sqrt{\sum_{i=1}^d\rho^2\left\|\left([X^T(t)]_i-\boldsymbol{1}\frac{\boldsymbol{1}^T[X^T(t)]_i}{|\LL|}\right)\right\|^2}\nonumber\\
&= \rho\left\|X(t)-\frac{X(t)\boldsymbol{1}}{|\LL|}\boldsymbol{1}^T\right\|_{\text{F}}.
\end{flalign}
It follows that
\begin{flalign}
&\|X(t+1)-\overline{X}(t+1)\|_{\text{F}}=\left\|X(t+1)-\frac{X(t+1)\boldsymbol{1}}{|\LL|}\boldsymbol{1}^T\right\|_{\text{F}}\nonumber\\
&=\left\|X(t)W^T(t)+\Delta(t)-\frac{[X(t)+\Delta(t)]\boldsymbol{1}}{|\LL|}\boldsymbol{1}^T\right\|_{\text{F}}\nonumber\\
&\stackrel{(a)}{\leq} \left\|X(t)W^T(t)-\frac{X(t)\boldsymbol{1}}{|\LL|}\boldsymbol{1}^T\right\|_{\text{F}}+\left\|\Delta(t)-\frac{\Delta(t)\boldsymbol{1}}{|\LL|}\boldsymbol{1}^T\right\|_{\text{F}}\nonumber\\
&\stackrel{(b)}{\leq}  \left\|X(t)W^T(t)-\frac{X(t)\boldsymbol{1}}{|\LL|}\boldsymbol{1}^T\right\|_{\text{F}}+\left\|\Delta(t)\right\|_{\text{F}}\nonumber\\
&\stackrel{(c)}{=} \left\|\left(X(t)-\frac{X(t)\boldsymbol{1}}{|\LL|}\boldsymbol{1}^T\right)W^T(t)\right\|_{\text{F}}+\left\|\Delta(t)\right\|_{\text{F}}\nonumber\\
&\stackrel{(d)}{\leq}\rho \left\|X(t)-\frac{X(t)\boldsymbol{1}}{|\LL|}\boldsymbol{1}^T\right\|_{\text{F}}+\left\|\Delta(t)\right\|_{\text{F}},
\end{flalign}
where $(a)$ follows from the triangle inequality, $(b)$ follows from \eqref{eq:delta_t_frob_upper}, $(c)$ follows from the double stochasticity of $W(t)$, and $(d)$ follows from \eqref{eq:upper_weight_prod_sterp_rho_upper}. 
Thus,
\begin{flalign*}
&\|X(t)-\overline{X}(t)\|\leq \rho^t\left\|X(0)-\frac{X(0)\boldsymbol{1}}{|\LL|}\boldsymbol{1}^T\right\|+\sum_{k=0}^{t-1} \rho^{t-1-k} \left\|\Delta(k)\right\|.
\end{flalign*}
Now, since $\EEop[\|\Delta_i(t)\|^2]\leq \delta^2(t)$ for all $i\in\LL$, and  by the non-negativity of the variance of $\left\|\Delta(k)\right\|_{\text{F}}$
\begin{flalign}
\EEop[\left\|\Delta(k)\right\|_{\text{F}}]
&\leq \sqrt{\EEop\left[\left\|\Delta(k)\right\|_{\text{F}}^2\right]}=\sqrt{\sum_{j\in\LL}\EEop\left[\|\Delta_j(k)\|^2\right]}\nonumber\\
&\leq\sqrt{|\LL|\delta^2(k)}=\sqrt{|\LL|}\delta(k),
\end{flalign}
it follows that
$\EEop[\|\Delta_i(t)\|]\leq \delta(t)$  for all $i\in\LL$. 

From \eqref{eq:max_val_norm} we have
\begin{flalign}
&\left\|X(0)-\frac{X(0)\boldsymbol{1}}{|\LL|}\boldsymbol{1}^T\right\|_{\text{F}}=\sqrt{\sum_{j\in\LL}\left\|\left[X(0)-\frac{X(0)\boldsymbol{1}}{|\LL|}\boldsymbol{1}^T\right]_j\right\|^2}\nonumber\\
&\leq\sqrt{\sum_{j\in\LL}\left(\left\|[X(0)]_j\right\|+\left\|\left[\frac{X(0)\boldsymbol{1}}{|\LL|}\boldsymbol{1}^T\right]_j\right\|\right)^2}\nonumber\\
&\leq\sqrt{\sum_{j\in\LL}\left(2\eta\right)^2}=2\eta\sqrt{|\LL|}.
\end{flalign}
It follows that
\begin{flalign}
&\frac{\EEop[\|X(t)-\overline{X}(t)\|_{\text{F}}]}{|\LL|}\leq \frac{2\eta}{\sqrt{|\LL|}}\rho^t+\frac{1}{\sqrt{|L|}}\sum_{k=0}^{t-1} \rho^{t-1-k} \delta(k)\nonumber\\
&\quad\leq \frac{2\eta}{\sqrt{|\LL|}}\rho^t+\frac{\delta(0)}{\sqrt{|L|}}\cdot\frac{\rho^{t/2}}{1-\rho}+\frac{\delta(t/2)}{\sqrt{|L|}(1-\rho)}.
\end{flalign}
Similarly,
\begin{flalign}
&\|X(t)-\overline{X}(t)\|_{\text{F}}^2\leq \rho^{2t}\left\|X(0)-\frac{X(0)\boldsymbol{1}}{|\LL|}\boldsymbol{1}^T\right\|^2\nonumber\\
&+2\rho^{t}\left\|X(0)-\frac{X(0)\boldsymbol{1}}{|\LL|}\boldsymbol{1}^T\right\|\sum_{k=0}^{t-1} \rho^{t-1-k}\left\|\Delta(k)\right\|_{\text{F}}\nonumber\\
&+\sum_{k_1=0}^{t-1}\sum_{k_2=0}^{t-1} \rho^{t-1-k_1}\rho^{t-1-k_2} \left\|\Delta(k_1)\right\|_{\text{F}}\cdot \left\|\Delta(k_2)\right\|_{\text{F}}\nonumber\\
&\leq \rho^{2t}4\eta^2|\LL|+4\rho^{t}\eta\sqrt{|\LL|}\sum_{k=1}^{t-1} \rho^{t-1-k}\left\|\Delta(k)\right\|_{\text{F}}\nonumber\\
&+\sum_{k_1=0}^{t-1}\sum_{k_2=0}^{t-1} \rho^{t-1-k_1}\rho^{t-1-k_2} \left\|\Delta(k_1)\right\|_{\text{F}}\cdot \left\|\Delta(k_2)\right\|_{\text{F}}.
\end{flalign}
By the Cauchy–Schwarz inequality \textit{for expectations}
\begin{flalign}
\EEop\left[\left\|\Delta(k_1)\right\|_{\text{F}}\cdot \left\|\Delta(k_2)\right\|_{\text{F}}\right]&\leq\sqrt{\EEop\left[\left\|\Delta(k_1)\right\|_{\text{F}}^2\right]\EEop\left[\left\|\Delta(k_2)\right\|_{\text{F}}^2\right]}\nonumber\\
&\leq |\LL|\delta(k_1)\delta(k_2).
\end{flalign}
Therefore,
\begin{flalign}
&\frac{\EEop\left[\|X(t)-\overline{X}(t)\|_{\text{F}}^2\right]}{|\LL|}\nonumber\\
&\leq \hspace{-0.05cm}4\eta^2\hspace{-0.05cm}\rho^{2t}\hspace{-0.05cm}+\hspace{-0.05cm}4\rho^{t}\eta\left[\frac{\delta(0)\rho^{t/2}}{1-\rho}\hspace{-0.05cm}+\hspace{-0.05cm}\frac{\delta(t/2)}{(1-\rho)}\right]\hspace{-0.05cm}+\hspace{-0.05cm}\left[\frac{\delta(0)\rho^{t/2}}{1-\rho}\hspace{-0.05cm}+\hspace{-0.05cm}\frac{\delta(t/2)}{(1-\rho)}\right]^2\nonumber\\
&=\left[2\eta\rho^{t}+\frac{\delta(0)\rho^{t/2}}{1-\rho}+\frac{\delta(t/2)}{(1-\rho)}\right]^2.
\end{flalign}
We  conclude the proof by utilizing \eqref{eq:upper_L21_frobenius2} and \eqref{eq:upper_L21_frobenius1}.
\end{proof}